\documentclass[journal,twoside,web]{ieeecolor}
\usepackage{generic}
\usepackage{cite}
\usepackage{amsmath,amssymb,amsfonts}
\usepackage{algorithmic}
\usepackage{graphicx}
\usepackage{algorithm,algorithmic}
\usepackage{hyperref}
\hypersetup{hidelinks=true}
\usepackage{textcomp}

\usepackage{algorithm}
\usepackage{subfigure}
\usepackage{multirow}
\usepackage{subcaption}
\usepackage{booktabs}
\usepackage{textcomp}
\usepackage{url}
\newtheorem{theorem}{Theorem}[section]
\newtheorem{condition}{Condition}[section]
\newtheorem{lemma}[theorem]{Lemma}
\newcommand{\ie}{\textit{i.e.}}
\def\BibTeX{{\rm B\kern-.05em{\sc i\kern-.025em b}\kern-.08em
    T\kern-.1667em\lower.7ex\hbox{E}\kern-.125emX}}
\begin{document}
\title{LadderMIL: Multiple-instance Learning with Coarse-to-Fine Self-Distillation}
\author{Shuyang Wu, Yifu Qiu, Ines P. Nearchou, Sandrine Prost, Jonathan A. Fallowfield, Hideki Ueno, \\Hitoshi Tsuda, David J. Harrison, Hakan Bilen, Timothy J. Kendall
\thanks{This work was supported by Medical Research Scotland under grant PHD-50538-2022.}
\thanks{Shuyang Wu, Sandrine Prost, Jonathan A. Fallowfield and Timothy J. Kendall are with the Centre for Inflammation Research, Institute for Regeneration and Repair, University of Edinburgh, Edinburgh, UK. (e-mail:frank.wu@ed.ac.uk; s.prost@ed.ac.uk; jonathan.fallowfield@ed.ac.uk; tim.kendall@ed.ac.uk)
}
\thanks{Yifu Qiu and Hakan Bilen are with the School of Informatics, University of Edinburgh, Edinburgh, UK. (e-mail: yifu.qiu@ed.ac.uk; h.bilen@ed.ac.uk)}
\thanks{Ines P. Nearchou is with the Indica Labs, 8700 Education Pl NW, Bldg. B Albuquerque, US. (e-mail: inearchou@indicalab.com)}
\thanks{David J. Harrison is with the School of Medicine, University of St Andrews, St Andrews, UK. (e-mail: david.harrison@st-andrews.ac.uk)}
\thanks{Hitoshi Tsuda is with the Pathology Centre and Department of Pathology, Chiba Medical Centre, Chiba, Japan. (e-mail: h-tsuda@seikei-kai.or.jp)}
\thanks{Hideki Ueno is with the Department of Surgery, National Defense Medical College, Tokorozawa, Japan. (e-mail: ueno\_surg1@ndmc.ac.jp)}
}

\maketitle

\begin{abstract}
Multiple Instance Learning (MIL) for whole slide image (WSI) analysis in computational pathology often neglects instance-level learning as supervision is typically provided only at the bag level, hindering the integrated consideration of instance and bag-level information during the analysis. In this work, we present \textbf{LadderMIL}, a framework designed to improve MIL through two perspectives: (1) \textit{employing instance-level supervision} and (2) \textit{learning inter-instance contextual information at bag level}. Firstly, we propose a novel Coarse-to-Fine Self-Distillation (\textbf{CFSD}) paradigm that probes and distils a network trained with bag-level information to adaptively obtain instance-level labels which could effectively provide the instance-level supervision for the same network in a self-improving way. Secondly, to capture inter-instance contextual information in WSI, we propose a Contextual Encoding Generator (\textbf{CEG}), which encodes the contextual appearance of instances within a bag. We also theoretically and empirically prove the instance-level learnability of CFSD. Our LadderMIL is evaluated on multiple clinically relevant benchmarking tasks including breast cancer receptor status classification, multi-class subtype classification, tumour classification, and prognosis prediction. By comparing with the best baseline, average improvements of 6.4\%, 5.4\% and 0.8\% in AUC, F1-score, and concordance index (C-index) are demonstrated across the five benchmarks; while average improvements of 7.6\% and 10.4\% in AUC and F1-score are shown in the external validation cohort. The code is available at: \url{https://github.com/franksyng/LadderMIL}.
\end{abstract}

\begin{IEEEkeywords}
Machine Learning, Computational Pathology, Multiple-instance Learning, Self-distillation
\end{IEEEkeywords}

\section{Introduction}
\label{sec:introduction}
Computational pathology (CPath) for the automated analysis of digital gigapixel whole slide images (WSIs) has demonstrated immense potential for precision medicine in fields typified by oncology~\cite{niazi2019e253,zhang2022tehm,khosravi2022pcr_bc,liang2023macronet,gao2024her2}. In contrast to regular daily images, WSIs pose two challenges~\cite{srinidhi2021cpath_survey}. 
Firstly, as expert annotation of features within an image is costly, WSIs are typically annotated with slide-level labels. Secondly, due to their extremely high resolution, it is a common requirement to divide WSIs into multiple patches and compute an embedding for each patch independently through a feature encoder before concatenating these embeddings into frozen bag-level features. Hence, the analysis of WSIs usually omits the online feature encoder, and patches in negatively labelled bags are assumed to all be negative while at least one is assumed to be positive in positive bags. Multiple instance learning (MIL)~\cite{dietterich1997mil} has been the standard machinery to model WSIs as a bag of patches (or instances) and to learn classification of them from only bag-level supervision.

\begin{figure*}[h]
\centering
% \begin{center}
\centerline{\includegraphics[width=\textwidth]{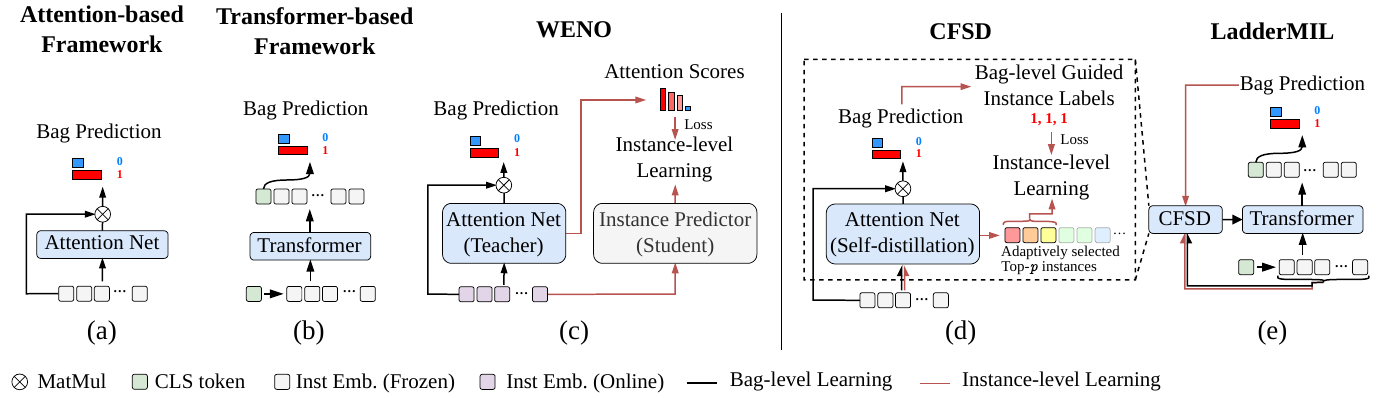}}
\caption{\textbf{Comparison of popular frameworks with our CFSD.} (a), (b) shows the traditional attention-based and transformer-based frameworks. (c) shows WENO~\cite{qu2022bidirectional}, a knowledge distillation framework that explicitly trains an instance predictor as student network. (d) illustrates our CFSD, a self-distillation framework that introduces instance-level learnability by taking one attention network to simultaneously learn knowledge from both bag-level and instance-level. (e) shows the brief illustration of our LadderMIL, a hybrid framework leveraging transformer and CFSD, capable of bag-level and instance-level learning in a self-improving way. The detailed overview of LadderMIL is in Figure~\ref{fig:overview}. Black arrows indicate bag-level learning while red arrows indicate instance-level learning.}
\label{fig:frameworks}
% \end{center}
\vskip -0.1in
\end{figure*}

Most conventional MIL frameworks in CPath are built on the success of deep networks. Due to the varying instance number, a pooling operation is commonly used in MIL to pool bag-level embeddings into a vector with fixed dimension. While maxpooling and average pooling are the most basic operations, the more successful techniques compute a weighted average of them by obtaining soft scores through attention mechanisms. Attention-based frameworks use gated attention mechanisms, such as ABMIL~\cite{ilse2018amil}, CLAM~\cite{lu_2021_data-efficient} and SCL-WC~\cite{wang2022scl}, where the latent features for classification are computed through matrix multiplying a bag-level attention map on bag-level features. Unlike these prior frameworks that consider instances independent to each other, vision transformers~\cite{dosovitskiy2021vit} have recently been applied to CPath problems to capture correlations across instances through multi-head attention~\cite{shao2021transmil, zhang2023attentionchallenging}, and DSMIL~\cite{li2021dsmil} introduced a distance-based dual-stream attention to model the instance correlation by training the distance measurement between each instance and the critical one with highest score. However, these prior approaches suffer from poor instance classification as learning to classify at bag level does not guarantee accurate learning at instance level due to the attention pooling operation that incorporates the hypothesis space for bag-level features into the instance predictions~\cite{jang2024learnability}. 

A promising strategy to provide instance-level supervision is knowledge distillation~\cite{hinton2015distillingknowledgeneuralnetwork}, which uses information from a teacher network (the bag classifier) to assist the training of a student network (the instance classifier). Self-distillation is a further simplified technique based on knowledge distillation that allows simultaneous knowledge sharing between the teacher and student networks. WENO~\cite{qu2022bidirectional} follows the knowledge distillation strategy to train bag and instance classifiers, using the bag-level soft pseudo labels to guide the instance-level training. However, although WENO trains the feature encoder from scratch with shared parameters between the two branches, it differs from the routine workflow in that a pre-trained backbone is applied to reduce computational costs~\cite{chen2022pancancer, kludt2024nextgen, he2024bladder, han2025hypergraph, ho2025f2fldm}. Although identifying high attention instances that are highly related to the prediction result is a crucial step for pseudo labelling and the instance-level learning, the selection mechanism in WENO remains inflexible and manually determined using grid search. As shown in Figure~\ref{fig:frameworks}, we compared our new method with three popular frameworks.

In this paper, to flexibly enable instance-level supervision for MIL, we introduce a novel Coarse-to-Fine Self-Distillation (CFSD) framework which facilitates learning from coarser (bag-level) knowledge to finer (instance-level) knowledge in a self-improving manner. In the bag-level branch, CFSD actively probes and distils the attention network trained with bag-level information to obtain instance-level labels for high-confidence instances. In the instance-level branch, the same attention network serves as an instance-level classifier and the selected high-confidence instances are used for further instance-level training. Unlike WENO, we show that powerful performance can be achieved by applying self-distillation directly on an attention network shared by the bag-level and instance-level branches, even when using frozen features that better align with the current application context. Additionally, an adaptive threshold scheduling (ATS) mechanism is designed to automatically update the threshold for high-attention instance selection during training, from the initial top-$5\%$ to a maximum of top-$20\%$ depending on whether the model continues to improve, offering more flexibility than a grid search approach.

Furthermore, we use the advantages of transformer-based frameworks that use self-attention~\cite{vaswani2017transformer} and positional encoding to capture inter-instance contextual information at the bag level. With the idea of conditional positional encoding~\cite{chu2023peg}, PEG~\cite{chu2023peg} and PPEG~\cite{shao2021transmil} gather information from neighbouring instances through reshaping feature sequences into square feature maps and applying convolutional operations. However, we argue that given WSIs vary in aspect ratio and many instances not adjacent in their original two-dimensional position are regarded as neighbours after background removal in preprocessing, the result of convolution is inaccurate. To address this, we propose the Contextual Encoding Generator (CEG) using intra-bag normalised $x$ and $y$ coordinates to provide accurate positional information incorporating with the attention map obtained from CFSD to more precisely encode the contextual arrangement of instances within a bag.

With the integration of these modules, we propose LadderMIL, a hybrid framework capable of bag-level and instance-level learning in a self-improving way, with CFSD and CEG plugged in. We demonstrate the efficacy of LadderMIL using five benchmarking tasks, including an internal benchmark for breast cancer estrogen and progesterone receptor status classification, multi-class subtype classification (TCGA-RCC\protect\footnote{https://www.cancer.gov/ccg/research/genome-sequencing/tcga}), tumour classification (CAMELYON16)~\cite{camelyon16}, and prognosis prediction (TCGA-LUAD). Our novel LadderMIL achieves the best performance across all benchmarks. Moreover, we theoretically prove the instance-level learnability of LadderMIL following Jang and Kwon~\cite{jang2024learnability}, and empirically validate it using the synthetic MNIST~\cite{deng2012mnist}, NCT-CRC-HE-100K NONORM~\cite{kather2018nct} and CAMELYON16 datasets.

Our main contributions are: (1) We propose CFSD and prove instance-level learnability both theoretically and empirically as a universal module that fits across various MIL frameworks. (2) We leverage a transformer-based framework and propose CEG for the encoding of inter-instance contextual information. (3) We show that LadderMIL, which integrates both CFSD and CEG, achieves state-of-the-art performance on multiple benchmarking tasks, introducing average improvements of 6.4\%, 5.4\% and 0.8\% in AUC, F1-score, and C-index, respectively.

\section{Related Work}
\subsection{Instance-level Learnability in MIL}
Recent studies have shown that the instance-level learnability of attention-based and transformer-based MIL models is not guaranteed, both theoretically and empirically~\cite{jang2024learnability}. This limitation arises from the attention pooling operation which multiplies attention weights over instance features, incorporating the hypothesis space for bag-level features into the instance predictions. While much effort has focused on improving MIL from the instance-level perspective, most work aims to either fine-tune feature extractors to obtain better representations~\cite{liu2023multiple, lin2023interventional, huang2024hnm, icmil2023, wang2024icmiltmi}, incorporate prior knowledge for specific disease types~\cite{li2026hcfmil}, or de-noise ambiguous instances for the bag-level learning~\cite{ren2023iibmil}. For instance, ICMIL~\cite{icmil2023, wang2024icmiltmi} iteratively couples the bag classifier and the patch encoder, using bag-level knowledge to fine-tune the encoder and regenerate improved instance representations. Such encoder-updating schemes require re-encoding the raw slides and adapting the feature encoder to the downstream data, while the MIL framework itself remains only at the bag level. These approaches are computationally expensive, which contradicts the goal of using MIL to reduce computational costs, especially with the availability of foundation models pre-trained on histopathological data~\cite{wang2021transpath, wang2022ctrans, xu2024gigapath, chen2024uni, vorontsov2024virchow}. HCF-MIL~\cite{li2026hcfmil} demonstrates a knowledge-driven end-to-end paradigm for the grading of laryngeal squamous cell carcinoma (LSCC), showing instance-level information is potentially significant for real-world clinical tasks. However, its knowledge signals require pathological guidance and the instance selection is based on clustering, where both need explicit human prior knowledge that reduces generalisable capabilities for tasks such as biomarker discovery and prognosis prediction. On the other hand, IIB-MIL~\cite{ren2023iibmil} introduces a label-disambiguation-based instance-level supervision using prototypes and confidence bank, but still serves to reduce the impact of noisy labels at bag-level training. Therefore, an efficient and self-improving approach to enable instance-level learning is needed to enhance MIL’s overall capability, and we consider self-distillation to be a possible option.

DTFD-MIL~\cite{zhang2022dtfd} employs feature distillation for two-tier bag-level training, creating smaller pseudo-bags to alleviate the effects of limited cohort sizes. Similarly, AHDMIL~\cite{dong2025hdmil, dong2025ahdmil} adopts a dual-branch mechanism that uses a Dynamic Multi-Instance Network (DMIN) for high resolution learning while distilling a Lightweight Instance Pre-screening Network (LIPN) to predict the relevance of each low-resolution patch. However, the training of DTFD-MIL and AHDMIL still focuses only on bag level. In contrast, WENO~\cite{qu2022bidirectional} uses knowledge distillation between the bag and instance levels, which takes the attention scores from positive instances at bag-level classification to be the soft pseudo labels that guide instance-level training. However, in WENO, the acquisition of positive instances relies on grid searching for the optimal threshold, which is inflexible since the positive instance ratio across different datasets usually varies. Furthermore, the parameter share in WENO is performed on the feature encoder and trained from scratch, whereas the previously mentioned pre-trained foundation models are now more widely used for feature extraction, omitting the update of feature encoder~\cite{kludt2024nextgen, he2024bladder, han2025hypergraph, ho2025f2fldm, jaume2024hest1k}. To solve these limitations, our CFSD is designed to train bag-level and instance-level classification on frozen features with a shared attention network, which uses self-distillation to improve the classifier instead of the feature encoder, and progressively introduces instance-level training by adaptively updating the threshold for high-attention instances selection, from top-$5\%$ to top-$20\%$.

\subsection{Positional Encoding in MIL}
TransMIL~\cite{shao2021transmil} performs MIL using two transformer layers with Nyström-Attention~\cite{xiong2021nystrom} and a Pyramid Position Encoding Generator (PPEG) to encode positional information. The idea of positional encoding for images originated in the Vision Transformer (ViT), which splits images into square patches and preserves all background patches as valid~\cite{dosovitskiy2021vit}. In standard images where all parts of the image are useful, this leads to only minor discontinuity between patches, except at row boundaries at the edge of the image. In contrast, when processing WSIs, non-informative and often abundant background patches without tissue are typically removed, creating significant discontinuities between the remaining patches containing informative tissue, introducing noise into positional encoding. Although PPEG resizes bag-level features into two dimensions and processes them with convolutions, the positional encoding is still one-dimensional and treats the embeddings as a continuous sequence. This disregards the spatial discontinuities, preventing PPEG from effectively representing the two-dimensional feature map and leading to inaccuracies in convolution operations. Instead, our CEG utilises normalised two-dimensional coordinates with the bag-level attention map obtained from CFSD to better capture the inter-instance relationships at the bag level.

\section{Methods}
This section describes the problem formulation, the design of CFSD, CEG and LadderMIL, and the approach used for evaluation.
\subsection{Multiple-instance Learning (MIL)}
\subsubsection{Problem formulation}
Taking binary classification as an example, given a bag of $K$ instances that $X=\{x_{1}, x_{2},...,x_{K}\}$, we would like to train a classifier that accurately predicts a bag-level target value $Y\in\{0,1\}$ without access to instance-level labels $\{y_{1}, y_{2},...,y_{K}\}$, where $y_{k}\in{\{0,1\}},k=1,2,...K$. The MIL problem is defined as:
\begin{equation}
Y =  
    \begin{cases}
    0, & \text{iff} \sum_{k} y_{k}=0, \\
    1, & \text{otherwise.}
    \end{cases}
\end{equation}

\subsubsection{Attention-based MIL} In computational pathology, a feature extractor with output dimension $1 \times D$ is used to create bag-level features $H=\{h_{1}, h_{2},...,h_{K}\}\in\mathbb{R}^{K\times D}$, where $h_{k}$ are instance-level embeddings. A fully connected layer is used as the first layer, reducing the embedding dimension to $512$, such that $\mathbf{h}\in\mathbb{R}^{K\times 512}$. To implement MIL with attention~\cite{ilse2018amil}, the attention network $f_{attn}$ comprises three linear layers with parameters $\mathbf{U}\in\mathbb{R}^{256\times 512}$, $\mathbf{V}\in\mathbb{R}^{256\times 512}$ and $\mathbf{w}\in\mathbb{R}^{256\times 1}$. The attention map $A_k\in \mathbb{R}^{K\times 1}$ and the attention-applied bag-level feature $\mathbf{M}\in \mathbb{R}^{1\times 512}$ are expressed as:
% \vspace{-3mm}
\begin{equation}
\label{eqn:gated_attn}
    A_{k} = \frac{\text{exp}\{\mathbf{w}^\top(\text{tanh}(\mathbf{V}\mathbf{h}_k^\top)\odot\text{sigm}(\mathbf{U}\mathbf{h}_k^\top))\}}{\sum_{j=1}^K\text{exp}\{\mathbf{w}^\top(\text{tanh}(\mathbf{V}\mathbf{h}_j^\top)\odot\text{sigm}(\mathbf{U}\mathbf{h}_j^\top))\}}
\end{equation}
\begin{equation}
\label{eqn:amil}
\mathbf{M}=\sum_{k=1}^{K}A_{k}\mathbf{h}_k
\end{equation}

\subsubsection{Transformer-based MIL} The transformer-based MIL framework differs from attention-based designs. Following the framework of TransMIL~\cite{shao2021transmil} that composes transformer layer $f_{NA}$ with $\text{Nyström-Attention}(\text{LN}(\cdot))$ and encodes position with PPEG, we construct our model using transformer layer $f_{SA}$ composed as $\text{Self-Attention}(\text{LN}(\cdot))$ and using the contextual encoding generator (CEG) for inter-instance contextual information encoding. Let $f_{cls}$ be the bag-level classifier, the classification made with TransMIL and our modified version are written as: 
% Let NA, SA, LN denote Nyström-Attention, Self-Attention and Layer Normalisation, the architecture of TransMIL $f_{trans}$ and our modified version $f'_{trans}$ are:
\begin{equation}
    \begin{aligned}
        \hat{Y}_{\text{TransMIL}} & =f_{cls}(f_{NA}(\text{PPEG}(f_{NA}(\cdot))))
    \end{aligned}
\end{equation}
\begin{equation}
\label{eqn:our_trans}
\begin{aligned}
    \hat{Y}_{\text{Ours}} &= f_{cls}(f_{SA}(\text{CEG}(f_{SA}(\cdot))))
\end{aligned}
\end{equation}

\subsection{Instance-level Learnable MIL}
\begin{figure}[h]
\begin{center}
\subfigure[Examples of top-$p$ importance instances.]{\centerline{\includegraphics[width=0.9\columnwidth]{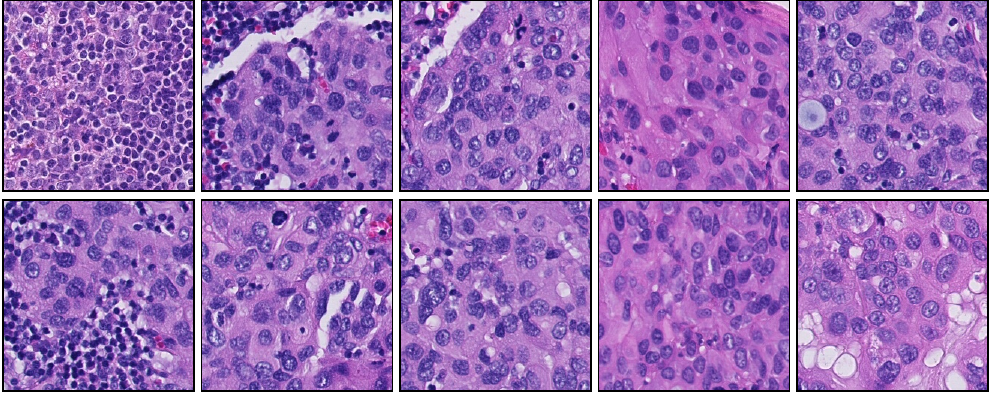}}}
\subfigure[Examples of reverse top-$p$ importance instances.]{\centerline{\includegraphics[width=0.9\columnwidth]{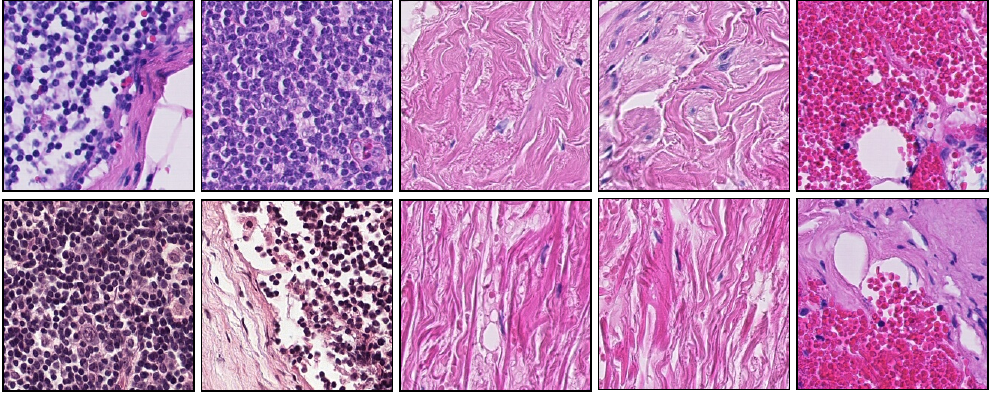}}}
\caption{\textbf{Instances visualisation of preliminary experiments.} In (a), we can see top-$p$ instances contain tumour areas, while in (b) the reverse top-$p$ instances contain mainly stroma, inflammatory cells, and red blood cells. The bag-level model can provide correct classification for top-$p$ instances.}
\label{fig:preliminary}
\end{center}
\end{figure}

% Before applying self-distillation on important instances derived from bag-level training, it is essential to ensure that the bag-level classifier focuses on the desired instances. Many
The attention-based framework has been widely used in previous work and it has been demonstrated that the attention network can highlight important instances related to the bag-level label~\cite{liang2023macronet, lu_2021_data-efficient, chen2022pancancer, huang2024pcr_rcc}, suggesting it is reasonable to annotate high-attention instances with bag-level labels and use self-distillation for instance-level supervision in MIL. We carried out a preliminary experiment to verify the principle. We applied CLAM-SB~\cite{lu_2021_data-efficient}, a model with attention-based framework, to the CAMEYLON16 benchmarking task. By comparing the instances with top-$p$ importance and reverse top-$p$ importance in the attention map $A$, we observed that CLAM-SB effectively focused on tumour instances (Figure~\ref{fig:preliminary}). This result suggests that annotating high-attention instances with bag-level labels is reasonable and highlights the potential for using self-distillation learning with bag-level knowledge, laying the groundwork for instance-level supervision in MIL. The implementation of the preliminary experiments follows the same protocol as the main experiments in Section~\ref{sec:implementations}.

\subsubsection{Coarse-to-Fine Self-Distillation (CFSD)}
Building on the previous work, we introduce the novel CFSD approach to improve instance-level learnability in MIL. CFSD probes and distils a network trained with bag-level information to adaptively obtain instance-level labels which could effectively provide the instance-level supervision for the same network in a self-improving way. Based on the finding that the top-$p$ instances are highly relevant to the prediction label $Y$, we apply an adaptive threshold scheduling (ATS) method which updates $p$ dynamically during training to select the top-$p$ instances $H'_{p}\in\mathbb{R}^{P\times D}$ and their corresponding instance-level label $Y'_{p}\in\{0,1\}$ from the bag $H$ using the trained attention map $A$ and bag-level label $Y$, where $p\in[5\%, 20\%]$. Initially, we set $p=5\%$ (top-$5\%$) to prioritise high-confidence instances, and the threshold $p$ is incremented by $1\%$ if the bag-level metrics (e.g., validation AUC/C-index) no longer increase for a patience of 3 consecutive epochs, to a maximum of $p=20\%$ (top-$20\%$). The threshold range of $p$ and the patience can be set in accordance to the user's needs. In this way, we ensure instance-level supervision is progressively introduced, providing flexibility and adaptability in the training. The pseudocode for self-annotating instance-level label is provided in Algorithm \ref{alg:1}.
\begin{algorithm}[H]
   \caption{\small Self-annotating instance-level label}
   \label{alg:1}
\begin{algorithmic}
   \STATE {\bfseries Input:} data $H$, target $Y$, attention map $A$
   \STATE {\bfseries Output:} inst\_data $H'$, inst\_target $Y'$
   \FOR{each bag}
   \STATE 1) argsort and rank attention score;
   \STATE $A_{\text{ranked}}\leftarrow \text{argsort}(A)/K$
   \STATE 2) get top-$p$ instances based on threshold $th$;
   \STATE $\text{selected\_idx}\leftarrow A_{\text{ranked}}>th\in [\text{top-}5\%,\text{top-}20\%]$
   \STATE 3) Get $H'$ and $Y'$;
   \STATE $H'\leftarrow H[\text{selected\_idx}]$,
   $Y'\leftarrow Y$ repeat for len(selected\_idx)
   \ENDFOR
   \STATE Given the bag number $m$:
   \STATE $H'_{all}\leftarrow\text{concat}(H'_1, H'_2,...,H'_m)$
   \STATE $Y'_{all}\leftarrow\text{concat}(Y'_1, Y'_2,...,Y'_m)$
\end{algorithmic}
\end{algorithm}

Once we have acquired the selected high attention instance-level embeddings across all bags, we concatenate them together to form all selected instances $H'_{all}$ and their corresponding labels $Y'_{all}$. Then, $H_{all}'$ and $Y'_{all}$ are used to regularly train the instance-level classifier. In the attention-based frameworks~\cite{ilse2018amil, lu_2021_data-efficient, li2021dsmil, wang2022scl, chen2022pathomic}, the attention network $f_{attn}$ can simultaneously act as the instance-level classifier since the output attention map $A\in\mathbb{R}^{K\times \mathcal{N}}$ can be interpreted as the classification of instances, where $\mathcal{N}$ denotes class number. Hence, in the instance-level branch, we optimise $f_{attn}$ instead of the bag-level classifier.
% Meanwhile, the implementation of CFSD in LadderMIL follows the same design such that $f_{attn}$ is applied before the transformer layers. Details will be introduced in section~\ref{sec:laddermil}.

The instance-level learnability of CFSD is theoretically proved following Jang and Kwon~\cite{jang2024learnability}. The hypothesis space for the $k^{th}$ instance is $\mathcal{H}_{inst_k}\cup\mathcal{H}_{add_k}$, where $\mathcal{H}_{inst_k}$ collects hypotheses depending only on $\mathbf{h}_k$ and $\mathcal{H}_{add_k}$ collects the additional hypotheses induced through elements outside the $k^{th}$ instance. Lemma~\ref{lemma:inst_learnability} establishes $\mathcal{H}_{add_k}\subset\mathcal{H}_{inst_k}$ (Condition~\ref{con:inst_learnability}) as a necessary condition for instance-level learnability, which holds when the instance prediction is generated from the instance's own features rather than from bag-level features.\\
\begin{proof}
Let $g(\cdot)$ denote the instance-level classifier of CFSD. In the instance-level branch, the shared attention network $f_{attn}$ acts directly as the instance classifier, which outputs the attention logit $A_{\text{logit},k}\in\mathbb{R}^{1\times\mathcal{N}}$ that being the per-instance classification:
\begin{equation*}
A_{\text{logit},k}=\mathbf{w}^\top(\text{tanh}(\mathbf{V}\mathbf{h}_k^\top)\odot\text{sigm}(\mathbf{U}\mathbf{h}_k^\top))
\end{equation*}
Since $\mathbf{U}$, $\mathbf{V}$, $\mathbf{w}$ can be regarded as acting on each independent instance without coupling instances, the instance decision function is as:
\begin{equation*}
G(h)=g_k(\mathbf{h}_k)
\end{equation*}
which depends solely on $\mathbf{h}_k$. Hence, CFSD supervises $f_{attn}$ directly rather than the pooled bag-level classifier, and the instance-level prediction carries no dependence on bag-level features.

Following Algorithm~\ref{alg:1}, a self-annotated label $Y'_k$ is applied specifically to an instance $\mathbf{h}_k$, instead of the overall bag-level. Therefore, each hypothesis for the $k^{th}$ instance is of the following form:
\begin{equation*}
\mathcal{H}_{add_k}=\{f_{\mathcal{H},k}:g_k(\mathbf{h}_k)\rightarrow Y'_k\}
\end{equation*}
which obeys the instance hypothesis space $\mathcal{H}_{inst_k}$ that produces results dependent solely on the $k^{th}$ instance feature:
\begin{equation*}
\mathcal{H}_{inst_k}=\{f_{\mathcal{H},k}:f_{\mathcal{H},k}(\mathbf{h}_k)\rightarrow Y'_k\}
\end{equation*}
Hence, Condition~\ref{con:inst_learnability} is satisfied that:
\begin{equation*}
\mathcal{H}_{add_k}\subset \mathcal{H}_{inst_k}
\end{equation*}
\end{proof}
and according to lemma~\ref{lemma:inst_learnability}, CFSD is instance-level learnable.

\subsubsection{Contextual Encoding Generator (CEG)}
\begin{figure}[h]
\centering
\includegraphics[width=0.45\textwidth]{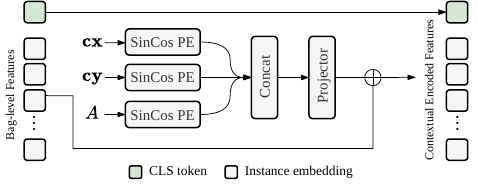}
\caption{\textbf{Overview of the Contextual Encoding Generator.} Normalised coordinates $(\mathbf{cx'},\mathbf{cy'})$ and attention map $A$ are encoded to obtain the contextual information.}
\label{fig:ceg}
\end{figure}
To mitigate the limitations caused by the discontinuity of instances in background-removed WSIs, we record the coordinates $(cx_k,cy_k)\in\mathbb{R}^{1\times 2}$ for each valid instance, and concatenate them to be coordinates in a bag, denoted as $(\mathbf{cx},\mathbf{cy})\in\mathbb{R}^{K\times 2}$. Given the aspect ratios of WSIs vary, the coordinates are normalised between 0 to 1 within each bag, such that $(\mathbf{cx'},\mathbf{cy'})=\{(cx'_1,cy'_1),...(cx'_k,cy'_k)\}$ with $cx_k',cy_k'\in [0,1]$. 
The $\mathbf{cx}$, $\mathbf{cy}$ and the attention map $A$ obtained from CFSD are together encoded to capture the contextual information:
\begin{equation}
    \begin{aligned}
        \mathbf{h}_{pe}= \mathbf{h} + \varphi(\text{concat}(\text{sincos($\mathbf{cx}'$)},\text{sincos($\mathbf{cy}'$)},\text{sincos($A$)}))
    \end{aligned}
\end{equation}

where $\mathbf{h}_{pe}$ denotes the encoded feature, and $\varphi$ is an MLP projector. The overview is shown in Figure~\ref{fig:ceg}.
% \begin{algorithm}[H]
%    \caption{\small Contextual Encoding Generator}
%    \label{alg:2}
% \begin{algorithmic}
%    \STATE {\bfseries Input:} data with CLS token $\mathbf{h}^\ell \in \mathbb{R}^{(k+1)\times 512}$, coordinates $(\mathbf{cx},\mathbf{cy})$, attention map $A$
%    \STATE {\bfseries Output:} context encoded embeddings $\mathbf{h}_{pe}^\ell$
%    \STATE 1) Normalise coordinates;
%    \FOR{each $(\mathbf{cx},\mathbf{cy})$}
%    \STATE $\text{max\_scale}\leftarrow max(max(\mathbf{cx}), max(\mathbf{cy}))$
%    \STATE $(\mathbf{cx}',\mathbf{cy}')\leftarrow
%    \text{min\_max\_scaler} (\mathbf{cx},\mathbf{cy}, \text{max\_scale})$
%    \ENDFOR
%    \STATE 2) Contextual encoding;
%    \STATE $\mathbf{h}, \mathbf{h}^{\ell(0)}\leftarrow \mathbf{h^\ell}$, where $\mathbf{h}^{\ell(0)}$ is the CLS token that $\mathbf{h}^{\ell(0)}\in\mathbb{R}^{1\times512}$
%    \STATE $\mathbf{h}_{pe}\leftarrow \mathbf{h} + \varphi(\text{concat}(\text{sincos($\mathbf{cx'}$)},\text{sincos($\mathbf{cy'}$)},\text{sincos($A$)}))$, where $\varphi$ is an MLP projector
%    \STATE $\mathbf{h}_{pe}^\ell\leftarrow \text{concat}(\mathbf{h}_{pe},\mathbf{h}^{\ell(0)})$
% \end{algorithmic}
% \end{algorithm}

\subsubsection{LadderMIL}
\label{sec:laddermil}

\begin{figure*}[h]
  \centering
  \begin{center}
  \centerline{\includegraphics[width=\textwidth]{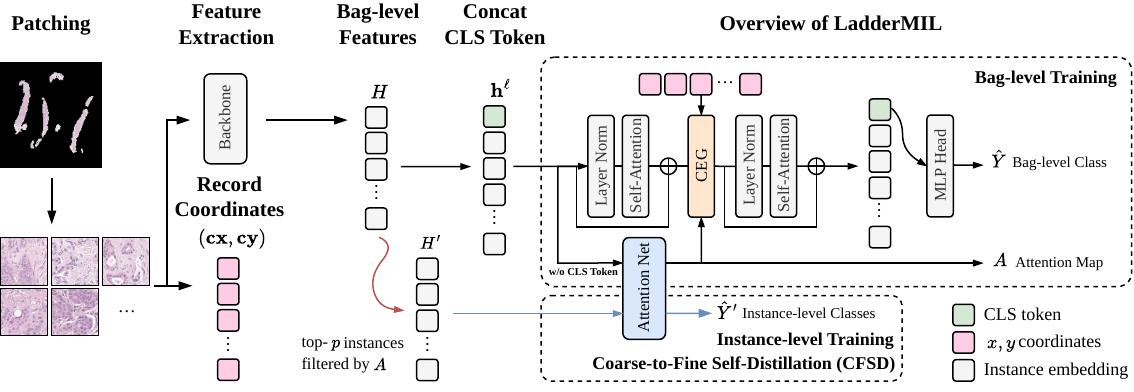}}
  \caption{\textbf{Overview of the LadderMIL.} CLAM~\cite{lu_2021_data-efficient} is used to remove the background and a pre-trained backbone is used to extract features from each patch. (1) The embedded bag-level features are then processed by the bag-level branch to obtain the bag-level prediction $\hat{Y}$ and attention map $A$. (2) Subsequently, the top-$p$ instances in each bag are selected and assigned a label according to the bag-level label, to form an instance-level dataset $H'$ with the corresponding labels $Y'$ across all bags. (3) Next, these data are used to train the instance-level branch. The (1), (2) and (3) steps are illustrated with black, red and blue arrows. In implementation, the bag-level network is firstly trained until it converged, and then applied CFSD to further train the bag-level and instance-level in a parallel way.}
  \label{fig:overview}
  \end{center}
\vskip -0.3in
\end{figure*}

LadderMIL is a hybrid framework with CFSD and CEG, as shown in Figure~\ref{fig:overview}. In the bag-level branch, we employ two transformer layers with self-attention and a CEG module located in between, and the CLS token is used for bag-level classification. Meanwhile, the bag level attention map $A$ is obtained from the attention network $f_{attn}$. Building upon Eqn(\ref{eqn:gated_attn})(\ref{eqn:amil})(\ref{eqn:our_trans}), and denoting the feature with CLS token as $\mathbf{h}^\ell$ and the bag-level prediction head as $f_{cls}$, the bag-level branch is composed as follow:
\begin{equation}
    \mathbf{h}=\text{FC}(H),\ \ \ \ 
% \end{equation}
% \begin{equation}
    A_\text{logit} =f_{attn}(\mathbf{h}),\ \ \ \
    A =\text{softmax}(A_\text{logit})\\
\end{equation}
\begin{equation}
    \hat{Y} = f_{cls}(f_{SA}(\text{CEG}(f_{SA}(\mathbf{h}, \mathbf{x}, \mathbf{y},A)))) \\
\end{equation}
% \begin{equation}
    % F_{bag} =f_{cls}(F_{trans}(\mathbf{h,\mathbf{x},\mathbf{y}},A))\\
% \end{equation}

while for classification tasks, the training is implemented following:
\begin{equation}
\mathcal{L}_{bag} = CELoss(\hat{Y}, Y)
\end{equation}

Note that our method can be also applied for prognosis prediction. Implemented following~\cite{chen2022pancancer, zadeh2021nllsurv}, the prognosis prediction is formulated as a four-class discrete-time survival modelling problem that splits patient survivorship into four discrete time slots. In preprocessing, to avoid data imbalance, data are distributed into four bins with equal cases number according to survival months. The annotation is made based on the bin that the case belongs to. Under this formulation, patients have vital status (caused death) are considered as uncensored while patients alive are censored. Let $Y_{hazard}$ and $Y_{surv}$ denote the predicted risk and survival rate, respectively, the censored loss $\mathcal{L}_{censored}$, uncensored loss $\mathcal{L}_{uncensored}$ and the loss for prognosis prediction $\mathcal{L}_{surv}$ are defined as follow:
\begin{equation}
Y_{hazard}=Sigmoid(F_{bag}(\mathbf{h},\mathbf{x},\mathbf{y}, A))
\end{equation}
\begin{equation}
Y_{surv}=\prod(1-Y_{hazard})
\end{equation}
\begin{equation}
\mathcal{L}_{censored}=-log(Y_{surv})
\end{equation}
\begin{equation}
\mathcal{L}_{uncensored}=-log(Y_{surv})-log(Y_{hazard})
\end{equation}
\begin{equation}
\mathcal{L}_{surv}=(1-\beta )\mathcal{L}_{censored}+\beta \mathcal{L}_{uncensored}
\end{equation}
, where $\beta$ is a variable for adjusting the weight of censored and uncensored loss.
    
For the instance-level branch, CFSD is applied to facilitate instance-level learning. The instance-level branch is written as follows:
\begin{equation}
        \hat{Y}'=f_{attn}(\mathbf{h}),\ \ \ \ 
\mathcal{L}_{inst} = CELoss(\hat{Y}', Y')
\end{equation}

By combining both branches, LadderMIL is trained by optimising the following objective:
\begin{equation}
\mathcal{L} = \mathcal{L}_{bag}+\mathcal{L}_{inst}
\end{equation}

The implementation of LadderMIL in pseudo-code is described in Algorithm \ref{alg:3}.
\begin{algorithm}[H]
   \caption{\small LadderMIL}
   \label{alg:3}
\begin{algorithmic}
   \STATE {\bfseries Input:} data $H$, coordinates $(\mathbf{cx},\mathbf{cy})$
   \FOR{each iteration}
   \STATE $\mathbf{h}\leftarrow \text{FC}(H)$, where $\mathbf{h}\in \mathbb{R}^{K\times 512}$
   \IF{bag-level}
   \STATE $A\leftarrow \text{softmax}(f_{attn}(\mathbf{h}))$
   \STATE $\mathbf{h}^\ell\leftarrow\text{concat}(\text{CLS}, \mathbf{h})$
   \STATE $\mathbf{h}'\leftarrow f_{SA}(\text{CEG}(f_{SA}(\mathbf{h^\ell}, \mathbf{x}, \mathbf{y},A)))$
   \STATE $\mathbf{h}''\leftarrow \text{layer\_norm}(\mathbf{h}')^{(0)}$, where $h''$ is the CLS token
   \STATE $\hat{Y}\leftarrow f_{cls}(h'')$
   \STATE {\bfseries Output:} $\hat{Y}$, $A$
   \ELSIF{instance-level}
   \STATE $\hat{Y}'\leftarrow f_{attn}(\mathbf{h})$
   \STATE {\bfseries Output:} $\hat{Y}'$
   \ENDIF
   \ENDFOR
\end{algorithmic}
\end{algorithm}

\section{Experiments}
\subsection{Bag-level experiments}
\subsubsection{Tasks and Datasets}
\label{sec:task_dataset}
We evaluate the performance of LadderMIL on five clinically relevant tasks.
\paragraph{Breast Cancer Receptor Status Classification.}
The receptor status of estrogen receptor (ER) and progesterone receptor (PR) inform treatment decision making, while the classification is challenging since not all tumour cells in a sample are guaranteed to be of the same receptor status due to tumour cell hormone receptor heterogeneity. We perform hormone receptor status classification on our internal breast cancer dataset which consists of 491 clinical cases reported by expert consultant breast pathologists. The internal data were supplied by the NHS Lothian BioResource (Research Tissue Bank ethical approval from East of Scotland Research Ethics Services ref: 25/ES/0030) with research project approval SR2036, 10 August 2023. The performance is further evaluated using an external cohort consisting of 232 cases. The external data were provided by the Department of Diagnostic Pathology at NDMC. The study was approved by the Ethics Committee of the NDMC, Tokorozawa, Japan (Approval Nos. 4913, 11 January 2024, and 5009, 16 May 2024). The annotation protocol is described in Appendix~\ref{appendix:receptor_annotation}.

\paragraph{Prognosis Prediction.} Prognosis prediction is a highly clinically relevant and challenging task that formulated as a four-class discrete-time survival modelling problem, which splits patient survivorship into four discrete time slots. We evaluate prognosis prediction performance on the TCGA-LUAD dataset which contains 465 cases with readily available follow-up clinical data including survival months and censorship.

\paragraph{Subtype Classification.} The capability of subtype classification is evaluated on the TCGA-RCC dataset, a kidney cancer dataset that contains three types of kidney cancer, including KIRC, KICH, and KIRP. After removing corrupted slides, the dataset consists of 919 diagnostic slides, with 517, 107, and 295 cases of the three subtypes, respectively.

\paragraph{Tumour Classification.} The capability of tumour classification is evaluated on the CAMELYON16 dataset, which is focused on tumour lymph node metastasis versus normal node classification in breast cancer. It consists of 270 training cases (160 normal and 110 tumour), and 130 test cases.

\subsubsection{Baseline Models}
To demonstrate the superior performance of our framework, we compared LadderMIL with several baseline models, including basic max-pooling and mean-pooling, ABMIL (attention-based pooling module)~\cite{ilse2018amil}, CLAM-SB and CLAM-MB~\cite{lu_2021_data-efficient}, AdditiveMIL~\cite{javed2022additive} and SCL-WC~\cite{wang2022scl} (gated attention), DSMIL (dual-stream MIL with instance and bag classifiers)~\cite{li2021dsmil} , TransMIL (PPEG and Nyström-Attention)~\cite{shao2021transmil}, DTFD-MIL (double-tier feature distillation)~\cite{zhang2022dtfd}, AHDMIL (asymmetric hierarchical distillation)~\cite{dong2025hdmil,dong2025ahdmil}, R$^2$T-MIL (feature re-embedding)~\cite{tang2024rrtmil}, IIB-MIL (label-disambiguation)~\cite{ren2023iibmil} and MFC-MIL (causality based overall data distribution estimation)~\cite{cui2025mfcmil}. Notably, the SimCLR~\cite{chen2020simclr}, a self-supervised contrastive learning method, was originally used to pre-train a ResNet-18 as the feature extractor for DSMIL. However, we omitted this step to mitigate the influence of data leakage caused by pre-training a feature encoder with downstream task data, and so as to better compare the robustness focusing on MIL frameworks.

Additionally, we specifically compared our CFSD with WENO~\cite{qu2022bidirectional} by evaluating the combination of ABMIL+CFSD and DSMIL+CFSD, then comparing the performance gaps with those of vanilla ABMIL and DSMIL. Then, we used these performance gaps to benchmark with the results of ABMIL+WENO and DSMIL+WENO, following their original paper.

\subsubsection{Implementations}
\label{sec:implementations}
\paragraph{Preprocessing}
We applied a consistent preprocessing protocol across all datasets, without data curation or normalisation, to better demonstrate our method's robustness to staining and scanning variation. The preprocessed data are applied to all compared methods in our experiments. WSIs were standardised to 0.2631 microns per pixel (MPP) and patched at $20\times$ magnification. Background removal and patching were performed using CLAM~\cite{lu_2021_data-efficient} and OpenSlide~\cite{goode2013openslide}, extracting non-overlapping patches of size $256\times256$.

\paragraph{Feature extraction}
We evaluated our method on features from two backbones. (1) Following the published prior work~\cite{wang2022ctrans, chen2022pancancer, javed2022additive}, we used an ImageNet pre-trained ResNet-50~\cite{he2016resnet} as a backbone, while embeddings were taken from the third layer, mean-pooled to obtain $1\times1024$ instance-level features, and concatenated to form the bag-level feature $H \in \mathbb{R}^{K\times1024}$. (2) To further evaluate the generalisability on different feature encoder, we also trained and evaluated on features extracted by specialised foundation model. We selected GigaPath~\cite{xu2024gigapath}, a foundation model pre-trained on histopathology data, since it is shown to perform significantly better among a series of foundation models on the majority of tasks in a previous benchmark~\cite{campanella2025benchmark}. The instance-level feature dimension for GigaPath is $1\times 1536$.

\paragraph{Experiment settings}
For evaluation, we used the area under the curve (AUC) and F1-score as performance metrics for classification tasks, while the concordance index (C-index) is used to measure prognosis prediction performance. All metrics are with scale of $[0,1]$ and the higher the better. We rigorously employed five-fold cross-validation for the training of all tasks. For our internal dataset, TCGA-RCC, and TCGA-LUAD datasets, we split the data into a train:val:test ratio of 3:1:1, reporting the average metrics on the test set. For the CAMELYON16 dataset, we divided the training data into a train:val ratio of 4:1 for five-fold cross-validation and evaluated the model on the official test set, with the average metrics from the test set reported.

\paragraph{Training details}
All experiments were undertaken on an RTX 3060 GPU. We used cross-entropy loss for both bag-level and instance-level training, with the AdamW optimiser~\cite{loshchilov2018adamw} and CosineAnnealing~\cite{loshchilov2017sgdr} scheduler for optimisation. The learning rate was set to $2\times 10^{-4}$ with batch size of $1$, while gradient accumulation was set to 32. We trained a total of 150 epochs, with early stopping applied if the metrics did not improve over 15 consecutive epochs. For fair comparison, we used the Lookahead optimiser~\cite{zhang2019lookahead} for TransMIL, adhering to their original design. For our LadderMIL, we first trained the bag-level network until it converged, and then applied CFSD to further train the bag-level and instance-level in a parallel way that bag-level branch is firstly fed forward and the instance-level branch follows in each epoch.

\subsection{Instance-level experiments}
We implement experiments to empirically prove the instance-level learnability on (1) a synthetic MNIST dataset~\cite{deng2012mnist} following Jang and Kwon~\cite{jang2024learnability}, (2) the real-world NCT-CRC-HE-100K NONORM~\cite{kather2018nct} and (3) CAMELYON16.
\subsubsection{Synthetic MNIST}
To demonstrate instance-level learnability, we followed the setup of Jang and Kwon~\cite{jang2024learnability} using a synthetic MNIST dataset. The task is framed as a multi-class classification MIL problem, with bag-level labels assigned as shown in Table~\ref{table:synthetic_mnist}. To isolate the impact of CEG and given that the MNIST dataset lacks inherent positional information, we employed CLAM-SB (baseline) with CFSD to assess instance-level learnability, rather than using LadderMIL with positional encoding. Hyperparameters were set in accordance with our main experiments, except for the learning rate, which was adjusted to 0.001. The MNIST dataset was split into 80\% training and 20\% evaluation data. Performance was evaluated using one-vs-rest AUC and F1-score. In this experiment, we not only empirically demonstrated instance-level learnability but also validated the multi-class classification capability of our framework.
\begin{table}[h]
\caption{\textbf{Annotation of the synthetic MNIST dataset.}}
\label{table:synthetic_mnist}
\begin{center}
\begin{small}
\begin{tabular}{c|l}
\toprule
Bag-label & Description \\
\midrule
\midrule
3   & the bag contains both 1 and 7\\
2   & the bag contains 1 but not 7\\
1   & the bag contains both 3 and 5\\
0   & other combinations\\
\bottomrule
\end{tabular}
\end{small}
\end{center}
\vskip -0.3in
\end{table}

\subsubsection{NCT-CRC-HE-100K NONORM}
We also show the instance-level learnability on real-world data, we evaluate our method on NCT-CRC-HE-100K NONORM, which contains 100,000 images in 9 tissue classes at 0.5 MPP~\cite{kather2018nct}. We created pseudo-slides following Table~\ref{table:nct_crc} for slide-level training, including 70\% of patches from TUM (tumour) and 90\% from other tissue types. BACK (background) is not included because in the current protocols, background tiles are generally removed. The remainder of the patches are used for the instance-level evaluation, testing the ability to classify TUM, LYM (lymph nodes), and others. The ImageNet pre-trained ResNet-50 is used for feature extraction while the bag-level training is implemented with the same hyper-parameters as in section~\ref{sec:implementations}.
\begin{table}[h]
\caption{\textbf{Annotation of the real-world NCT-CRC-HE-100K NONORM dataset.}}
\label{table:nct_crc}
\begin{center}
\begin{small}
\begin{tabular}{c|l}
\toprule
Bag-label & Description \\
\midrule
\midrule
2   & others+LYM+TUM\\
1   & others+LYM\\
0   & others\\
\bottomrule
\end{tabular}
\end{small}
\end{center}
% \vskip -0.1in
\end{table}

\subsubsection{CAMELYON16}
Furthermore, we also demonstrate the instance-level learnability over the real world task of tumour/non-tumour classification over CAMELYON16 dataset. The task is formulated as binary classification that aligns with the bag-level task. The ImageNet pre-trained ResNet-50 is used for feature extraction while the bag-level training is implemented with the same hyper-parameters as in section~\ref{sec:implementations}.

\begin{table*}[h]
\centering
% \vskip -0.2in
\caption{\textbf{Model comparison on ResNet-50 extracted features.}
% Evaluated with ER+/ER- classification and PR+/PR- classification on the internal dataset, subtype classification on TCGA-RCC, tumour/normal classification on CAMELYON16 and prognosis prediction on TCGA-LUAD.
\textbf{Bold} indicates overall the best while \underline{underline} indicates the best in subgroup. The standard deviation is determined with 2000 bootstrap trails.}
% \vskip 0.1in
\label{table:model_compare_rn50}
\resizebox{\textwidth}{!}{
\begin{tabular}{l|ccccccccc}
\toprule
\multirow{2}{*}{Dataset \& Metrics} & \multicolumn{2}{c}{Internal (ER)} & \multicolumn{2}{c}{Internal (PR)} & \multicolumn{2}{c}{TCGA-RCC}      & \multicolumn{2}{c}{CAMELYON16}    & TCGA-LUAD       \\
\cmidrule{2-10}
                                    & AUC             & F1-score        & AUC             & F1-score        & AUC             & F1-score        & AUC             & F1-score        & C-index         \\
\midrule
\midrule
MeanPooling                         & 64.58$_{\pm3.56}$          & 55.18$_{\pm1.91}$          & 64.82$_{\pm2.90}$          & 57.80$_{\pm2.08}$          & 95.58$_{\pm0.56}$          & 80.40$_{\pm1.66}$          & 61.94$_{\pm2.24}$          & 56.50$_{\pm2.23}$          & 54.00$_{\pm1.69}$          \\
MaxPooling                          & 65.60$_{\pm3.29}$          & 54.90$_{\pm1.91}$          & 64.88$_{\pm2.27}$          & 54.74$_{\pm2.21}$          & 96.50$_{\pm0.51}$          & 85.02$_{\pm1.44}$          & 67.56$_{\pm2.31}$          & 60.50$_{\pm2.27}$          & 49.64$_{\pm1.85}$          \\
ABMIL                               & 65.10$_{\pm3.67}$          & 55.72$_{\pm2.11}$          & 60.78$_{\pm2.85}$          & 54.58$_{\pm2.20}$          & 97.50$_{\pm0.39}$          & 84.30$_{\pm1.51}$          & 62.06$_{\pm2.29}$          & 58.38$_{\pm2.26}$          & 59.52$_{\pm1.77}$          \\
CLAM-SB                             & 86.58$_{\pm2.36}$          & 72.02$_{\pm1.32}$          & 66.56$_{\pm2.95}$          & 60.14$_{\pm2.19}$          & 98.38$_{\pm0.31}$          & 89.44$_{\pm1.27}$          & 75.52$_{\pm2.06}$          & 65.92$_{\pm2.15}$          & 53.96$_{\pm1.76}$          \\
CLAM-MB                             & 83.70$_{\pm2.67}$          & 69.46$_{\pm1.36}$          & 70.70$_{\pm2.64}$          & 60.92$_{\pm1.91}$          & 98.42$_{\pm0.31}$          & 88.94$_{\pm1.31}$          & 72.66$_{\pm2.22}$          & 65.00$_{\pm2.22}$          & 52.98$_{\pm1.72}$          \\
DSMIL                               & 67.94$_{\pm3.50}$          & 55.76$_{\pm1.75}$          & 62.04$_{\pm2.91}$          & 59.82$_{\pm2.14}$          & 97.16$_{\pm0.43}$          & 85.96$_{\pm1.39}$          & 63.42$_{\pm2.34}$          & 60.46$_{\pm2.23}$          & 57.06$_{\pm1.78}$          \\
TransMIL                            & 66.06$_{\pm3.38}$          & 55.76$_{\pm2.01}$          & 61.42$_{\pm2.86}$          & 55.44$_{\pm2.31}$          & 98.50$_{\pm0.29}$          & 88.60$_{\pm1.26}$          & 62.54$_{\pm2.24}$          & 56.76$_{\pm2.29}$          & 50.50$_{\pm1.63}$          \\
AdditiveMIL                         & 86.02$_{\pm2.36}$         & 72.52$_{\pm1.33}$          & 69.36$_{\pm2.77}$          & 63.00$_{\pm1.96}$          & 98.40$_{\pm0.30}$          & 89.48$_{\pm1.29}$          & 73.90$_{\pm2.12}$          & 64.10$_{\pm2.19}$          & 53.96$_{\pm1.76}$          \\
SCL-WC                              & 84.42$_{\pm2.79}$          & 69.72$_{\pm1.38}$          & 76.16$_{\pm2.45}$          & 66.20$_{\pm1.79}$          & 98.28$_{\pm0.32}$          & 89.20$_{\pm1.30}$          & 71.04$_{\pm2.13}$          & 64.28$_{\pm2.13}$          & 57.04$_{\pm1.77}$          \\
DTFD-MIL                              & 70.54$_{\pm3.45}$          & 59.76$_{\pm2.06}$          & 67.70$_{\pm2.75}$          & 61.54$_{\pm2.21}$          & 95.60$_{\pm0.84}$          & 83.36$_{\pm1.62}$          & 72.74$_{\pm1.92}$          & 67.94$_{\pm2.18}$          & 52.78$_{\pm1.66}$          \\
AHDMIL & \underline{87.90}$_{\pm2.41}$ & \underline{72.74}$_{\pm1.45}$ & \underline{79.62}$_{\pm2.54}$ & \underline{71.14}$_{\pm1.97}$ & \underline{98.70}$_{\pm0.27}$ & \underline{90.56}$_{\pm1.25}$ & \underline{75.84}$_{\pm2.02}$ & 71.82$_{\pm2.20}$ & \underline{60.46}$_{\pm2.60}$ \\
IIB-MIL & 73.94$_{\pm3.73}$ & 59.88$_{\pm2.05}$ & 68.60$_{\pm2.62}$ & 65.72$_{\pm2.10}$ & 98.40$_{\pm0.31}$ & 88.92$_{\pm1.32}$ & 65.92$_{\pm2.28}$ & 61.42$_{\pm2.31}$ & 56.76$_{\pm2.52}$ \\
R$^2$T-MIL & 83.88$_{\pm3.35}$ & 71.24$_{\pm1.63}$ & 72.06$_{\pm2.52}$ & 63.20$_{\pm2.01}$ & 98.56$_{\pm0.33}$ & 88.66$_{\pm1.22}$ & 75.42$_{\pm2.05}$ & \underline{73.96}$_{\pm2.15}$ & 59.84$_{\pm2.55}$ \\
MFC-MIL & 75.98$_{\pm3.38}$ & 62.68$_{\pm1.88}$ & 71.42$_{\pm2.58}$ & 63.70$_{\pm1.93}$ & 97.28$_{\pm0.40}$ & 86.16$_{\pm1.39}$ & 66.78$_{\pm2.28}$ & 65.20$_{\pm2.33}$ & 54.42$_{\pm2.75}$ \\

\midrule
\textbf{LadderMIL (Ours)}         & \textbf{91.78}$_{\pm1.96}$ & \textbf{78.48}$_{\pm1.14}$ & \textbf{84.72}$_{\pm2.26}$ & \textbf{75.90}$_{\pm1.72}$ & \textbf{99.24}$_{\pm0.22}$          & \textbf{93.02}$_{\pm1.05}$          & \textbf{86.54}$_{\pm1.79}$ & \textbf{77.22}$_{\pm1.88}$ & \textbf{60.96}$_{\pm1.85}$\\
\bottomrule
\end{tabular}
}
\end{table*}

\begin{table*}[h]
\centering
\caption{\textbf{Model comparison on GigaPath extracted features.} \textbf{Bold} indicates overall the best while \underline{underline} indicates the best in subgroup. The standard deviation is determined with 2000 bootstrap trails.}
\label{table:model_compare_gigapath}
\resizebox{\textwidth}{!}{
\begin{tabular}{l|ccccccccc}
\toprule
\multirow{2}{*}{Dataset \& Metrics} & \multicolumn{2}{c}{Internal (ER)} & \multicolumn{2}{c}{Internal (PR)} & \multicolumn{2}{c}{TCGA-RCC}      & \multicolumn{2}{c}{CAMELYON16}    & TCGA-LUAD       \\
\cmidrule{2-10}
                                    & AUC             & F1-score        & AUC             & F1-score        & AUC             & F1-score        & AUC             & F1-score        & C-index         \\
\midrule
\midrule
MeanPooling                         & 86.36$_{\pm2.63}$          & 70.40$_{\pm1.37}$          & 82.64$_{\pm2.20}$          & 72.96$_{\pm1.64}$       & 82.64$_{\pm0.25}$          & 72.96$_{\pm1.04}$        & 82.64$_{\pm2.37}$          & 72.96$_{\pm2.26}$         & 58.46$_{\pm1.80}$          \\
MaxPooling                          & 85.78$_{\pm2.57}$          & 66.66$_{\pm1.48}$          & 76.52$_{\pm2.52}$          & 67.98$_{\pm1.77}$  &  98.82$_{\pm0.27}$  &  91.68$_{\pm1.12}$ &   67.92$_{\pm2.14}$  &    61.32$_{\pm2.20}$     & 49.78$_{\pm1.75}$          \\
ABMIL                               & 92.16$_{\pm1.79}$          & 79.26$_{\pm1.13}$          & 84.52$_{\pm2.06}$          & \underline{75.72}$_{\pm1.55}$   & 99.18$_{\pm0.2}$ & 92.78$_{\pm1.07}$ & 94.32$_{\pm0.82}$ & 90.88$_{\pm1.46}$       & 55.58$_{\pm1.78}$          \\
CLAM-SB                             & 92.72$_{\pm1.71}$          & 77.52$_{\pm1.23}$          & 85.00$_{\pm1.97}$          & 73.10$_{\pm1.52}$   & \underline{99.26}$_{\pm0.19}$ & 93.72$_{\pm0.95}$ & 97.58$_{\pm0.69}$ & 93.98$_{\pm1.15}$       & 59.40$_{\pm1.77}$          \\
CLAM-MB                             & \underline{92.80}$_{\pm1.78}$          & \underline{80.30}$_{\pm1.22}$          & \underline{85.72}$_{\pm1.86}$          & 74.94$_{\pm1.55}$    & 99.22$_{\pm0.21}$ & 94.20$_{\pm0.94}$ & 97.20$_{\pm0.69}$ & 92.48$_{\pm1.38}$      & 56.38$_{\pm1.73}$          \\
DSMIL                               & 90.08$_{\pm1.98}$          & 76.36$_{\pm1.18}$          & 84.18$_{\pm2.18}$          & 75.24$_{\pm1.57}$    & 99.18$_{\pm0.21}$ & 92.72$_{\pm1.05}$ & 66.70$_{\pm2.29}$ & 61.34$_{\pm2.28}$      & 59.94$_{\pm1.75}$          \\
TransMIL                            & 88.76$_{\pm2.23}$          & 73.86$_{\pm1.15}$          & 83.18$_{\pm2.22}$          & 73.54$_{\pm1.59}$    & 99.10$_{\pm0.23}$ & 93.50$_{\pm0.96}$ & 62.86$_{\pm2.37}$ & 57.44$_{\pm2.32}$      & 60.30$_{\pm1.81}$          \\
AdditiveMIL                         & 92.82$_{\pm1.74}$          & 77.96$_{\pm1.27}$          & 85.04$_{\pm1.97}$          & 74.12$_{\pm1.52}$    & 99.17$_{\pm0.21}$ & 93.94$_{\pm0.96}$ & 97.98$_{\pm0.67}$ & 93.46$_{\pm1.19}$      & 59.46$_{\pm1.77}$          \\
SCL-WC                              & 92.20$_{\pm1.70}$          & 78.68$_{\pm1.16}$          & 85.30$_{\pm1.89}$          & 73.76$_{\pm1.51}$    & 99.24$_{\pm0.21}$ & \underline{94.88}$_{\pm0.87}$ & 96.94$_{\pm0.73}$ & 92.96$_{\pm1.33}$      & \underline{61.08}$_{\pm1.81}$          \\
DTFD-MIL                            & 78.86$_{\pm3.71}$          & 70.48$_{\pm1.92}$          & 81.54$_{\pm2.94}$          & 72.52$_{\pm2.14}$    & 99.02$_{\pm0.21}$ & 91.24$_{\pm1.19}$ & 87.78$_{\pm1.14}$ & 84.98$_{\pm1.80}$      & 53.12$_{\pm1.67}$          \\
AHDMIL & 88.26$_{\pm2.14}$ & 75.52$_{\pm1.17}$ & 84.98$_{\pm2.07}$ & 75.74$_{\pm1.55}$ & 98.84$_{\pm0.24}$ & 91.60$_{\pm1.16}$ & 75.54$_{\pm1.97}$ & 70.16$_{\pm2.10}$ & 50.16$_{\pm2.54}$ \\
IIB-MIL & 91.76$_{\pm2.51}$ & 77.34$_{\pm1.30}$ & 81.36$_{\pm2.17}$ & 74.46$_{\pm1.58}$ & 99.00$_{\pm0.36}$ & 91.58$_{\pm1.14}$ & \underline{98.46}$_{\pm0.59}$ & \underline{94.16}$_{\pm1.15}$ & 60.26$_{\pm2.55}$ \\
R$^2$T-MIL & 89.80$_{\pm2.18}$ & 73.56$_{\pm1.24}$ & 84.26$_{\pm2.05}$ & 74.16$_{\pm1.56}$ & 99.04$_{\pm0.23}$ & 92.60$_{\pm1.07}$ & 95.58$_{\pm1.01}$ & 91.84$_{\pm1.51}$ & 59.20$_{\pm2.54}$ \\
MFC-MIL & 88.00$_{\pm2.01}$ & 72.76$_{\pm1.32}$ & 82.62$_{\pm2.41}$ & 74.46$_{\pm1.77}$ & 98.52$_{\pm0.34}$ & 89.74$_{\pm1.09}$ & 84.24$_{\pm1.81}$ & 80.42$_{\pm2.00}$ & 55.38$_{\pm2.37}$ \\

\midrule
% Our LadderMIL$^{\dagger}$            & 93.54          & 80.14          & 84.70          & 75.58          & 60.82          \\
\textbf{LadderMIL (Ours)}       & \textbf{95.22}$_{\pm1.79}$ & \textbf{81.86}$_{\pm1.07}$ & \textbf{86.44}$_{\pm2.07}$ & \textbf{76.12}$_{\pm1.54}$ & \textbf{99.42}$_{\pm0.23}$ & \textbf{95.96}$_{\pm0.76}$ & \textbf{98.84}$_{\pm0.48}$ & \textbf{95.08}$_{\pm1.09}$ & \textbf{63.32}$_{\pm1.75}$\\
\bottomrule
\end{tabular}
}
\end{table*}

\section{Results}
In this section, the model performance comparison, ablation studies on each module, and model interpretability are shown and discussed. Metrics are reported in percentage.
\subsection{Model comparisons}
% \subsubsection{Compared on ResNet-50 extracted features}
% \subsubsection{Compared on ResNet-50 extracted features}
The models trained with ResNet-50 extracted features are compared in Table \ref{table:model_compare_rn50}. LadderMIL achieved significant performance improvements over the baseline models on all benchmarks, with AUC scores of 91.78 and 84.72 for ER and PR receptor status classification, respectively, an AUC of 99.34 for subtype classification, an AUC of 86.54 for tumour classification, and a C-index of 60.96 for prognosis prediction. On average, LadderMIL obtained improvements of 6.4\%, 5.4\%, and 0.8\% in AUC, F1-score, and C-index, respectively, across the five benchmarks compared to the best baseline.

Among the baselines, AHDMIL with dual-branch handling both high resolution analysis and lightweight instance pre-screening for correlation learning stably outperformed the others. In contrast, DSMIL showed limited performance due to the absence of SimCLR pre-trained features, highlighting a lack of robustness. Similarly, TransMIL underperformed in the training scheme using Lookahead optimiser that followed their original design, even compared to models without positional encoding. We attribute this to the susceptibility of PPEG to instance discontinuity, hindering its ability to capture true contextual relationships.

% \subsubsection{Compared on GigaPath extracted features}
Additionally, models trained on GigaPath-extracted features were evaluated and compared in Table~\ref{table:model_compare_gigapath}. Our LadderMIL continued to outperform other baselines, achieving up to 3\% improvement in prognosis prediction, further demonstrating its generalisability.

% \subsubsection{CFSD vs. WENO}
We also benchmarked our CFSD with the other knowledge distillation method \ie, WENO. In the comparison between WENO and CFSD, we focus on the performance gap rather than the absolute performance, in order to mitigate the influence of differences in data splitting and hyperparameter settings. As shown in Table~\ref{table:cfsd_vs_weno}, CFSD significantly outperformed WENO on both ResNet-50 and GigaPath features, achieving the highest AUC improvements of 14.86 and 28.88 for ABMIL and DSMIL, respectively.
\begin{table}[h]
\centering
\caption{\textbf{Comparing CFSD and WENO on CAMELYON16.} The performance gap $\Delta$ versus the vanilla model in AUC score is reported.}
\label{table:cfsd_vs_weno}
% \resizebox{0.35\textwidth}{!}{
\begin{tabular}{l|cc|cc}
\toprule
Models & \multicolumn{2}{c|}{ABMIL}         & \multicolumn{2}{c}{DSMIL}         \\
\midrule
Metric & \multicolumn{2}{c|}{AUC} & \multicolumn{2}{c}{AUC}    \\
% \cmidrule{2-5}
\midrule
Features           & ResNet-50   & GigaPath   & ResNet-50  & GigaPath   \\
\midrule
\midrule
$\times$                 & 62.06$_{\pm2.29}$          & 94.32$_{\pm0.82}$          & 63.42$_{\pm2.34}$          & 66.70$_{\pm2.29}$           \\
WENO                 & 66.88$_{\pm2.24}$          & 93.30$_{\pm0.90}$          & 73.18$_{\pm1.89}$          & 93.84$_{\pm0.77}$           \\
% CFSD (Parallel)          & \textbf{81.62} & 96.60          & 74.36          & 90.16          \\
CFSD       & \textbf{76.92}$_{\pm1.95}$          & \textbf{97.74}$_{\pm0.69}$ & \textbf{75.62}$_{\pm2.15}$ & \textbf{95.58}$_{\pm1.09}$ \\
\midrule
% $\Delta$CFSD (Parallel)    & \textbf{\color{red}+19.56} & \color{red}+2.28          & \color{red}+10.94          & \color{red}+23.46          \\
$\Delta$CFSD & \textbf{\color{red}+14.86}          & \textbf{\color{red}+3.42} & \textbf{\color{red}+12.20} & \textbf{\color{red}+28.88} \\
\midrule
$\Delta$WENO               & {\color{red}+4.82}   & {\color{blue}-1.02}    & {\color{red}+9.76}   &  {\color{red}+27.14}  \\
\bottomrule
\end{tabular}
\vskip -0.2in
% }
\end{table}

% \subsubsection{External evaluation}
To further demonstrate the generalisability of LadderMIL on unseen data, we undertook model comparison for ER and PR status classification on an external cohort in Table~\ref{table:external}. It is shown that the attention-based frameworks, namely the ABMIL, CLAM, AdditiveMIL and SCL-WC generalise better in the external evaluation, while performance degradation is witnessed on AHDMIL. We show that LadderMIL has good generalisability and consistently outperforms other baselines in the external cohort, with average improvements of 7.6\%, 10.4\% in AUC and F1-score over the best baseline on ResNet-50 features.
\begin{table*}[h]
\centering
\vskip -0.2in
\caption{\textbf{Model comparison on the external cohort for ER and PR classification.} \textbf{Bold} indicates overall the best while \underline{underline} indicates the best in subgroup. The standard deviation is determined with 2000 bootstrap trails.}
\label{table:external}
\resizebox{0.8\textwidth}{!}{
\begin{tabular}{l|cccc|cccc}
\toprule
 & \multicolumn{4}{c|}{ResNet50 Features} & \multicolumn{4}{c}{GigaPath Features} \\
\cmidrule{2-9}
Framework & \multicolumn{2}{c}{External (ER)} & \multicolumn{2}{c|}{External (PR)} & \multicolumn{2}{c}{External (ER)} & \multicolumn{2}{c}{External (PR)} \\
\cmidrule{2-9}
 & AUC & F1 & AUC & F1 & AUC & F1 & AUC & F1 \\
\midrule
MeanPooling   & 55.40$_{\pm2.73}$ & 52.02$_{\pm1.91}$ & 58.62$_{\pm2.18}$ & 54.54$_{\pm2.08}$ & 77.72$_{\pm2.28}$ & 60.92$_{\pm1.37}$ & 74.72$_{\pm1.78}$ & 62.56$_{\pm1.64}$ \\
MaxPooling    & 63.88$_{\pm2.61}$ & 54.78$_{\pm1.91}$ & 62.50$_{\pm2.08}$ & 52.52$_{\pm2.11}$ & 76.86$_{\pm2.28}$ & 63.24$_{\pm1.48}$ & 72.38$_{\pm1.92}$ & 57.28$_{\pm1.77}$ \\
ABMIL         & 56.50$_{\pm2.74}$ & 49.40$_{\pm2.11}$ & 60.04$_{\pm2.03}$ & 54.68$_{\pm2.20}$ & \underline{85.14}$_{\pm1.83}$ & 68.86$_{\pm1.13}$ & 77.66$_{\pm1.74}$ & 64.52$_{\pm1.55}$ \\
CLAM-SB       & \underline{79.54}$_{\pm2.13}$ & 61.60$_{\pm1.32}$ & 68.70$_{\pm2.11}$ & 56.38$_{\pm2.19}$ & 84.96$_{\pm1.83}$ & 67.54$_{\pm1.23}$ & 78.08$_{\pm1.78}$ & 67.08$_{\pm1.52}$ \\
CLAM-MB       & 75.64$_{\pm2.31}$ & 59.28$_{\pm1.36}$ & 70.06$_{\pm1.96}$ & 58.94$_{\pm1.91}$ & 84.80$_{\pm1.95}$ & 67.04$_{\pm1.22}$ & \underline{78.74}$_{\pm1.77}$ & \underline{68.38}$_{\pm1.55}$ \\
DSMIL         & 58.70$_{\pm2.68}$ & 55.36$_{\pm1.75}$ & 58.34$_{\pm2.20}$ & 51.12$_{\pm2.14}$ & 82.76$_{\pm2.05}$ & 64.20$_{\pm1.18}$ & 77.10$_{\pm1.78}$ & 66.32$_{\pm1.57}$ \\
TransMIL      & 62.10$_{\pm2.59}$ & 54.90$_{\pm2.01}$ & 55.30$_{\pm2.20}$ & 48.90$_{\pm2.31}$ & 81.20$_{\pm2.13}$ & 65.80$_{\pm1.15}$ & 73.90$_{\pm1.96}$ & 62.54$_{\pm1.59}$ \\
AdditiveMIL   & 78.82$_{\pm2.18}$ & \underline{62.60}$_{\pm1.33}$ & 70.72$_{\pm2.06}$ & 56.00$_{\pm1.96}$ & 84.96$_{\pm1.83}$ & 67.54$_{\pm1.27}$ & 78.08$_{\pm1.75}$ & 67.08$_{\pm1.52}$ \\
SCL-WC        & 76.10$_{\pm2.26}$ & 58.56$_{\pm1.38}$ & \underline{73.24}$_{\pm1.79}$ & \underline{62.44}$_{\pm1.79}$ & 84.80$_{\pm1,75}$ & \underline{69.48}$_{\pm1.16}$ & 78.18$_{\pm1.79}$ & 67.60$_{\pm1.51}$ \\
DTFD-MIL      & 61.92$_{\pm2.61}$ & 54.30$_{\pm2.06}$ & 65.12$_{\pm2.05}$ & 58.96$_{\pm2.21}$ & 79.78$_{\pm2.42}$ & 63.92$_{\pm1.92}$ & 76.46$_{\pm2.19}$ & 61.02$_{\pm2.14}$ \\
AHDMIL & 74.77$_{\pm5.09}$ & 60.94$_{\pm2.21}$ & 71.30$_{\pm3.55}$ & 60.05$_{\pm2.18}$ & 66.64$_{\pm4.27}$ & 53.83$_{\pm2.29}$ & 67.15$_{\pm3.95}$ & 57.39$_{\pm2.31}$ \\
IIB-MIL & 61.12$_{\pm6.07}$ & 54.88$_{\pm2.90}$ & 62.10$_{\pm4.34}$ & 57.73$_{\pm2.77}$ & 81.89$_{\pm4.79}$ & 68.13$_{\pm1.75}$ & 76.81$_{\pm3.93}$ & 66.82$_{\pm2.07}$ \\
R$^2$T-MIL & 76.32$_{\pm5.02}$ & 59.19$_{\pm1.90}$ & 68.12$_{\pm3.71}$ & 58.32$_{\pm2.26}$ & 80.40$_{\pm4.60}$ & 65.30$_{\pm1.99}$ & 74.85$_{\pm3.90}$ & 66.56$_{\pm2.28}$ \\
MFC-MIL & 70.19$_{\pm5.00}$ & 49.66$_{\pm2.52}$ & 64.81$_{\pm4.28}$ & 54.75$_{\pm2.44}$ & 79.36$_{\pm4.89}$ & 64.45$_{\pm2.26}$ & 70.99$_{\pm4.29}$ & 61.62$_{\pm2.31}$ \\
\midrule
\textbf{LadderMIL (Ours)} & \textbf{82.58}$_{\pm2.31}$ & \textbf{68.48}$_{\pm1.14}$ & \textbf{81.62}$_{\pm1.71}$ & \textbf{68.44}$_{\pm1.72}$ & \textbf{86.06}$_{\pm2.21}$ & \textbf{70.86}$_{\pm1.07}$ & \textbf{82.08}$_{\pm1.84}$ & \textbf{69.92}$_{\pm1.54}$ \\
\bottomrule
\end{tabular}
}
\end{table*}

\subsection{Ablation study}
\subsubsection{Efficacy of CFSD and CEG}
We next assessed the effectiveness of CFSD and CEG in Table~\ref{table:ablation_rn50},~\ref{table:ablation_gigapath}. The results demonstrate that CFSD leads to obvious improvements in the attention-based framework of CLAM-SB. This highlights that enhancing learnability at the instance level is empirically beneficial to bag-level learning. Additionally, we show that the efficacy of CEG is substantial. LadderMIL with CEG outperforms the combinations with other positional encoding modules measured by both AUC, F1-score, and C-index, including PPEG. This improvement is attributed to the encoding of the accurate coordinates with the bag-level attention map, which better captures inter-instance contextual information in background-removed WSI.
\begin{table*}[h]
\centering
\caption{\textbf{Ablation study of CFSD and CEG on ResNet-50 extracted features.} \textbf{Bold} indicates overall the best while \underline{underline} indicates the best in subgroup. The standard deviation is determined with 2000 bootstrap trails.}
% \vskip 0.1in
\label{table:ablation_rn50}
\resizebox{\textwidth}{!}{
\begin{tabular}{l|cc|ccccccccc}
\toprule
\multirow{2}{*}{Framework} & \multicolumn{2}{c|}{Modules}                         & \multicolumn{2}{c}{Internal (ER)} & \multicolumn{2}{c}{Internal (PR)} & \multicolumn{2}{c}{TCGA-RCC}      & \multicolumn{2}{c}{CAMELYON16}    & TCGA-LUAD       \\
\cmidrule{2-12}
                    & CFSD        & PE   & AUC             & F1-score        & AUC             & F1-score        & AUC             & F1-score        & AUC             & F1-score        & C-index         \\
\midrule
\midrule
CLAM-SB                        & $\times$           & $\times$    & 86.58$_{\pm2.36}$          & 72.02$_{\pm1.32}$          & 66.56$_{\pm2.95}$          & 60.14$_{\pm2.19}$          & 98.38$_{\pm0.31}$          & 89.44$_{\pm1.27}$          & 75.52$_{\pm2.06}$          & 65.92$_{\pm2.15}$          & 53.96$_{\pm1.76}$          \\
% CLAM-SB                        & Parallel    & $\times$    & 86.80          & 71.04          & 78.56          & 67.90          & 98.58          & 88.40          & \underline{86.20}    & \underline{77.16}    & 57.90          \\
CLAM-SB                        & \checkmark & $\times$    & \underline{86.88}$_{\pm2.37}$    & \underline{73.60}$_{\pm1.31}$    & \underline{81.50}$_{\pm2.49}$    & \underline{70.64}$_{\pm1.86}$    & \underline{98.80}$_{\pm0.39}$    & \underline{90.94}$_{\pm1.13}$    & \underline{84.72}$_{\pm2.17}$          & \underline{75.92}$_{\pm2.06}$          & \underline{59.96}$_{\pm1.65}$    \\
\midrule
\multirow{7}{*}{LadderMIL (Ours)} & $\times$           & Random   & 85.88$_{\pm2.38}$          & 69.82$_{\pm1.39}$          & 71.86$_{\pm2.51}$          & 68.00$_{\pm1.90}$          & 98.80$_{\pm0.27}$          & 89.66$_{\pm1.20}$          & 80.56$_{\pm1.76}$          & 71.62$_{\pm2.04}$          & 57.16$_{\pm1.67}$          \\

& $\times$           & 1D   & 88.58$_{\pm2.54}$          & 73.38$_{\pm1.39}$          & 63.62$_{\pm2.86}$          & 56.86$_{\pm2.07}$          & 98.70$_{\pm0.28}$          & 90.60$_{\pm1.26}$          & 84.48$_{\pm1.77}$          & 75.22$_{\pm2.07}$          & 51.94$_{\pm1.75}$          \\
                               & $\times$           & 2D   & 89.06$_{\pm2.57}$          & 76.56$_{\pm1.49}$          & 80.38$_{\pm2.52}$          & 69.98$_{\pm1.87}$          & 98.64$_{\pm0.28}$          & 89.50$_{\pm1.25}$          & 78.66$_{\pm2.09}$          & 69.48$_{\pm2.26}$          & 59.08$_{\pm1.74}$          \\
                               & $\times$           & PPEG & 89.62$_{\pm2.15}$          & 75.96$_{\pm1.28}$          & 77.56$_{\pm2.44}$          & 68.86$_{\pm1.84}$          & 98.68$_{\pm0.26}$          & 89.78$_{\pm1.24}$          & 82.60$_{\pm1.72}$          & 73.64$_{\pm2.05}$          & 58.88$_{\pm1.73}$          \\
                               & \checkmark           & 2D & 91.36$_{\pm2.35}$          & 77.70$_{\pm1.60}$          & 84.32$_{\pm2.41}$          & 74.24$_{\pm1.78}$          & 98.82$_{\pm0.26}$          & 92.24$_{\pm1.08}$          & 85.94$_{\pm1.73}$          & 76.84$_{\pm2.06}$          & 59.74$_{\pm1.83}$          \\
                               % & Parallel    & PPEG & 90.78          & 77.56          & 82.58          & 72.72          & 99.22          & 92.86          & 81.96          & 73.64          & 59.64          \\
                               & \checkmark & PPEG & 90.56$_{\pm2.03}$          & 77.24$_{\pm1.23}$          & 82.74$_{\pm2.45}$          & 73.20$_{\pm1.75}$          & 99.20$_{\pm0.23}$          & 92.70$_{\pm1.00}$          & 85.88$_{\pm1.68}$          & 75.82$_{\pm1.96}$          & 60.92$_{\pm1.79}$          \\
                               % & Parallel    & CEG  & 91.34          & 78.00          & 84.54          & 74.54          & \textbf{99.34} & \textbf{93.38} & 84.76          & 76.52          & 59.82          \\
                               & \checkmark & CEG  & \textbf{91.78}$_{\pm1.96}$ & \textbf{78.48}$_{\pm1.14}$ & \textbf{84.72}$_{\pm2.26}$ & \textbf{75.90}$_{\pm1.72}$ & \textbf{99.24}$_{\pm0.22}$          & \textbf{93.02}$_{\pm1.05}$          & \textbf{86.54}$_{\pm1.79}$ & \textbf{77.22}$_{\pm1.88}$ & \textbf{60.96}$_{\pm1.85}$\\
\bottomrule
\end{tabular}
}
\end{table*}

\begin{table*}[h]
\centering
\small
\caption{\textbf{Ablation study of CFSD and CEG on GigaPath extracted features.} \textbf{Bold} indicates overall the best while \underline{underline} indicates the best in subgroup. The standard deviation is determined with 2000 bootstrap trails.}
% \vskip 0.1in
\label{table:ablation_gigapath}
\resizebox{\textwidth}{!}{
\begin{tabular}{l|cc|ccccccccc}
\toprule
\multirow{2}{*}{Framework} & \multicolumn{2}{c|}{Modules}                         & \multicolumn{2}{c}{Internal (ER)} & \multicolumn{2}{c}{Internal (PR)} & \multicolumn{2}{c}{TCGA-RCC}      & \multicolumn{2}{c}{CAMELYON16}    & TCGA-LUAD       \\
\cmidrule{2-12}
                    & CFSD        & PE   & AUC             & F1-score        & AUC             & F1-score        & AUC             & F1-score        & AUC             & F1-score        & C-index         \\
\midrule
\midrule
CLAM-SB                        & $\times$                & $\times$        & 92.72$_{\pm1.71}$          & 77.52$_{\pm1.23}$          & 85.00$_{\pm1.97}$          & 73.10$_{\pm1.52}$   & 99.26$_{\pm0.19}$ & 93.72$_{\pm0.95}$ & 97.58$_{\pm0.69}$ & 93.98$_{\pm1.15}$      & 59.40$_{\pm1.77}$          \\
% CLAM-SB                        & Parallel         & $\times$        & 93.84          & 80.02          & 85.32          & \underline{75.74}    & 55.12          \\
CLAM-SB                        & \checkmark      & $\times$        & \underline{94.38}$_{\pm1.53}$    & \underline{80.80}$_{\pm1.15}$    & \underline{85.50}$_{\pm2.01}$  & \underline{74.56}$_{\pm1.63}$  & \underline{99.36}$_{\pm0.18}$ & \underline{95.62}$_{\pm0.74}$ & \underline{97.80}$_{\pm0.74}$ & \underline{94.26}$_{\pm1.34}$        & \underline{62.80}$_{\pm1.82}$    \\
\midrule
\multirow{7}{*}{LadderMIL (Ours)} & $\times$                & Random       & 91.06$_{\pm1.86}$          & 75.14$_{\pm1.30}$          & 85.30$_{\pm1.97}$          & 75.16$_{\pm1.62}$  & 98.98$_{\pm0.32}$ & 93.68$_{\pm1.01}$ & 98.10$_{\pm0.64}$ & 94.58$_{\pm1.40}$       & 59.96$_{\pm1.76}$          \\
& $\times$                & 1D       & 93.18$_{\pm1.97}$          & 80.90$_{\pm1.23}$          & 85.10$_{\pm2.06}$          & 72.16$_{\pm1.57}$  &  99.06$_{\pm0.26}$	& 94.62$_{\pm0.90}$	& 98.08$_{\pm0.50}$ & 94.80$_{\pm1.10}$    & 54.66$_{\pm1.64}$          \\
                               & $\times$                & 2D       & 93.08$_{\pm1.88}$          & 79.78$_{\pm1.23}$          & 85.50$_{\pm2.11}$          & 75.90$_{\pm1.58}$  & 99.12$_{\pm0.31}$	& 94.34$_{\pm0.90}$	& 98.28$_{\pm0.52}$	& 94.44$_{\pm1.13}$        & 61.62$_{\pm1.84}$          \\
                               & $\times$                & PPEG     & 90.16$_{\pm2.03}$          & 77.98$_{\pm1.40}$          & 85.10$_{\pm2.17}$          & 74.38$_{\pm1.71}$   & 99.20$_{\pm0.25}$	& 95.04$_{\pm0.83}$	& 98.00$_{\pm0.51}$	& 94.60$_{\pm1.03}$       & 57.94$_{\pm1.74}$          \\
                               % & Parallel         & PPEG     & 93.16          & 79.74          & 85.86          & 76.06          & 59.54          \\
                                & \checkmark      & 2D     & 94.40$_{\pm2.35}$          & 80.72$_{\pm1.35}$          & 86.02$_{\pm2.09}$          & 75.84$_{\pm1.63}$   & 99.20$_{\pm0.19}$	& 94.86$_{\pm0.86}$	& 98.42$_{\pm0.52}$	& 94.00$_{\pm1.27}$       &        62.56$_{\pm1.70}$   \\
                               & \checkmark      & PPEG     & 94.58$_{\pm1.43}$          & 80.86$_{\pm1.13}$          & 85.90$_{\pm1.92}$          & 75.78$_{\pm1.54}$    & 99.22$_{\pm0.20}$	& 95.68$_{\pm0.79}$	& 98.62$_{\pm0.57}$	& 94.76$_{\pm1.16}$      & 62.64$_{\pm1.67}$          \\
                               % & Parallel         & CEG      & 93.54          & 80.14          & 84.70          & 75.58          & 60.82          \\
                               & \checkmark      & CEG     & \textbf{95.22}$_{\pm1.79}$ & \textbf{81.86}$_{\pm1.07}$ & \textbf{86.44}$_{\pm2.07}$ & \textbf{76.12}$_{\pm1.54}$ & \textbf{99.42}$_{\pm0.23}$ & \textbf{95.96}$_{\pm0.76}$ & \textbf{98.84}$_{\pm0.48}$ & \textbf{95.08}$_{\pm1.09}$ & \textbf{63.32}$_{\pm1.75}$\\
\bottomrule
\end{tabular}
}
\end{table*}

\subsubsection{Efficacy of ATS}
The performance of LadderMIL with fixed top-$p$ settings, including top-$5\%$, top-$10\%$, top-$15\%$ and top-$20\%$, was compared with the ATS applied counterparts in receptor classification. It is shown in Table~\ref{table:diffk} that ATS succeeded in flexibly adjusting the top-$p$ threshold during training and introduced better performance.
\begin{table}[h]
\centering
\caption{\textbf{Comparison of fixed threshold selection with Adaptive Threshold Scheduling.} The standard deviation is determined with 2000 bootstrap trails.}
\label{table:diffk}
\resizebox{\columnwidth}{!}{
\begin{tabular}{l|l|cccc}
\toprule
\multirow{2}{*}{Model}                       & \multirow{2}{*}{Threshold} & \multicolumn{2}{c}{Internal (ER)} & \multicolumn{2}{c}{Internal (PR)}  \\
\cmidrule{3-6}
                                             &                            & AUC             & F1-score        & AUC             & F1-score        \\
\midrule
\midrule
% \multirow{4}{*}{Our LadderMIL (Parallel)}    & Top-5\%                    & 90.12          & 77.88          & 84.20          & 74.12          \\
%                                              & Top-10\%                   & 90.80          & 77.98          & 84.22          & 74.42          \\
%                                              & Top-15\%                   & 90.50          & 77.06          & 84.08          & 74.08          \\
%                                              & ATS                        & \underline{91.34}    & \underline{78.00}    & \underline{84.54}    & \underline{74.54}    \\
% \midrule
\multirow{4}{*}{LadderMIL (Ours)} & Top-5\%                    & 91.18$_{\pm2.06}$          & 78.22$_{\pm1.33}$          & 84.58$_{\pm2.31}$          & 75.64$_{\pm1.77}$          \\
                                             & Top-10\%                   & 91.30$_{\pm2.26}$          & 78.02$_{\pm1.51}$          & 84.34$_{\pm2.21}$          & 74.00$_{\pm1.72}$          \\
                                             & Top-15\%                   & 91.54$_{\pm2.09}$          & 74.08$_{\pm1.39}$          & 84.24$_{\pm2.37}$          & 74.76$_{\pm1.71}$          \\
                                             & Top-20\%                   & 89.02$_{\pm3.61}$          & 76.10$_{\pm1.85}$          & 78.46$_{\pm2.82}$          & 71.96$_{\pm2.05}$          \\
                                             & ATS                        & \textbf{91.78}$_{\pm1.96}$ & \textbf{78.48}$_{\pm1.14}$ & \textbf{84.72}$_{\pm2.26}$ & \textbf{75.90}$_{\pm1.72}$ \\
\bottomrule
\end{tabular}
}
\vskip -0.2in
\end{table}

\subsubsection{Empirical proof of instance-level learnability}
We demonstrate instance-level learnability using (1) the synthetic MNIST dataset~\cite{deng2012mnist}, following the approach outlined by Jang and Kwon~\cite{jang2024learnability} and (2) the real-world histology image dataset NCT-CRC-HE-100K NONORM~\cite{kather2018nct}. As shown in Table \ref{table:inst_learnability}, the CLAM-SB baseline shows limited performance in instance-level classification on both the synthetic MNIST dataset and the real-world histology image dataset. In contrast, CFSD is empirically proven to enable instance-level learning.
\begin{table}[h]
% \small
\centering
\caption{\textbf{Comparison of bag-level performance $P_{bag}$ and instance-level $P_{inst}$ performance for CFSD on the synthetic MNIST dataset, NCT-CRC-HE-100K NONORM and CAMELYON16.} The framework of CLAM-SB (baseline) and the CFSD plugged-in counterparts are tested in the experiment.}
\label{table:inst_learnability}
\resizebox{0.45\textwidth}{!}{
\begin{tabular}{l|cc|cc|cc}
\toprule
\multirow{2}{*}{Inst. Learning}     & \multicolumn{2}{c|}{$P_{bag}$}          & \multicolumn{2}{c|}{$P_{inst}$}         & \multicolumn{2}{c}{$P_{inst}-P_{bag}$}      \\
\cmidrule{2-7}
        & AUC             & F1-score        & AUC             & F1-score        & AUC              & F1-score         \\
\midrule
\midrule
\multicolumn{7}{c}{MNIST} \\
\midrule
$\times$           & 92.40          & 73.46          & 47.55          & 6.59          & -44.85          & -66.87          \\
WENO          & 92.14          & 73.72          & 79.78          & 35.21          & -12.36          & -38.51          \\
CFSD & \textbf{92.54}          & \textbf{75.50}          & \textbf{86.45}          & \textbf{40.16}          & \textbf{-6.09}          & \textbf{-35.34}         \\
\midrule
\midrule
\multicolumn{7}{c}{NCT-CRC-HE-100K NONORM} \\
\midrule
$\times$           & 99.70          & 100          & 33.76          & 9.94          & -65.94          & -90.06          \\
WENO          & 100          & 100          & 53.74          & 33.83          & -46.26          & -66.17          \\
CFSD & \textbf{100}          & \textbf{100}          & \textbf{60.45}          & \textbf{36.47}          & \textbf{-39.55}          & \textbf{-63.53}         \\
\midrule
\midrule
\multicolumn{7}{c}{CAMELYON16} \\
\midrule
$\times$           & 90.50          & 88.28          & 71.70          & 67.68          & -18.80          & -20.60          \\
WENO          & 94.90          & 92.76          & 78.98          & 72.64          & -15.92          & -20.12          \\
CFSD & \textbf{95.28}          & \textbf{94.12}          & \textbf{94.02}          & \textbf{88.81}          & \textbf{-1.26}          & \textbf{-5.31}         \\
\bottomrule
\end{tabular}
}
\end{table}

\subsubsection{Training efficiency}
We also analysed the training efficiency of LadderMIL versus other baselines. The complexity is mainly $\mathcal{O}((1+p)KD^2)$, where $K$ is the number of instances within a bag, $p$ is the top-$p$ percentage and $D$ denotes the feature dimension. The training time of LadderMIL on bag-level only and CFSD activated is shown in Table~\ref{table:efficiency_epoch_time}. The bag-level training time cost of IIB-MIL and TransMIL is approximating the LadderMIL with top-$15\%$ and top-$20\%$, respectively. Therefore, LadderMIL provides instance-level supervision in an affordable time cost as training a larger scale bag-level model. Meanwhile, since the instance-level training only starts after the bag-level training is converged, we avoid the time cost of training both levels from scratch. Furthermore, the instance-level branch is only activated during training, while in the inference stage only the bag-level branch is used for whole-slide analysis. For epoch-wise inference time (Table~\ref{table:efficiency_epoch_spd_brief}), IIB-MIL and TransMIL explicitly take a longer time in each epoch than LadderMIL. Combining the aforementioned results, LadderMIL brings distinct performance improvements versus other models, while limiting the maximum epoch-wise inference time gap to only around 0.57s.

\begin{table}[h]
\centering
\caption{\textbf{Training time compared on the ER classification training set with 289 cases.} Median time/epoch in seconds among 15 runs on an RTX 3060 GPU is reported.}
\label{table:efficiency_epoch_time}
% \resizebox{0.35\textwidth}{!}{
% \begin{tabular}{l|c|c|c|c}
% \toprule
% Model & CFSD     & $\times$                     & \checkmark   & Patch Num. \\
% \midrule
% \multirow{4}{*}{LadderMIL} & Top-5\%  & \multirow{4}{*}{6.16} & 11.97 & 31,616     \\
% & Top-10\% &                       & 14.83 & 63,389     \\
% & Top-15\% &                       & 16.95 & 95,128     \\
% & Top-20\% &                       & 18.96 & 126,918   \\
% \midrule
% TransMIL & $\times$ &     15.19   & $\times$ & $\times$   \\
% DTFD-MIL & $\times$ &     11.23   & $\times$ & $\times$   \\
% ABMIL & $\times$ &     4.73   & $\times$ & $\times$   \\
% CLAM-SB & $\times$ &     5.15   & $\times$ & $\times$   \\
% CLAM-MB & $\times$ &     5.68   & $\times$ & $\times$   \\
% AdditiveMIL & $\times$ &     5.30   & $\times$ & $\times$   \\
% SCL-WC & $\times$ &     6.17   & $\times$ & $\times$   \\
% DSMIL & $\times$ &     5.97   & $\times$ & $\times$   \\
% \bottomrule
% \end{tabular}
\begin{tabular}{l|c|c|c|c}
\toprule
Model & Bag only & \multicolumn{2}{c|}{CFSD (bag + inst.)} & Patch Num. \\
\midrule
\multirow{4}{*}{LadderMIL} & \multirow{4}{*}{2.24} & Top-5\% & 5.15 & 31,616 \\
& & Top-10\% & 6.75 & 63,389 \\
& & Top-15\% & 8.31 & 95,128 \\
& & Top-20\% & 9.83 & 126,918 \\
\midrule
ABMIL        & 0.46  & \multirow{12}{*}{$\times$} &  \multirow{12}{*}{$\times$}  & \multirow{12}{*}{$\times$}\\
CLAM-SB      & 0.81  & &  \\
CLAM-MB      & 0.92  & &  \\
DSMIL        & 0.60  & &  \\
TransMIL     & 10.31 & &  \\
AdditiveMIL  & 0.82  & &  \\
SCL-WC       & 0.95  & &  \\
DTFD-MIL     & 1.70  & &  \\
AHDMIL       & 2.33  & &  \\
R$^2$T-MIL       & 2.85  & &  \\
MFC-MIL       & 4.41  & &  \\
IIB-MIL       & 7.90  & &  \\
\bottomrule
\end{tabular}
% }
\vskip -0.2in
\end{table}

\begin{table}[h]
\centering
\caption{\textbf{Bag-level inference time compared on the ER classification training set with 289 cases.} Median time/epoch in seconds among 15 runs on an RTX 3060 GPU is reported.}
\label{table:efficiency_epoch_spd_brief}
\resizebox{0.35\textwidth}{!}{
\begin{tabular}{l|c|c}
\toprule
Models & Time/Epoch (s)      & Gap vs. Ours \\
\midrule
\midrule
MeanPooling  & 0.299 & \textcolor{blue}{-0.57} \\
MaxPooling   & 0.700 & \textcolor{blue}{-0.17} \\
ABMIL        & 0.337 & \textcolor{blue}{-0.53} \\
CLAM-SB      & 0.434 & \textcolor{blue}{-0.43} \\
CLAM-MB      & 0.498 & \textcolor{blue}{-0.37} \\
DSMIL        & 0.389 & \textcolor{blue}{-0.48} \\
TransMIL     & 3.301 & \textcolor{red}{+2.44} \\
AdditiveMIL  & 0.438 & \textcolor{blue}{-0.43} \\
SCL-WC       & 0.506 & \textcolor{blue}{-0.36} \\
DTFD-MIL     & 0.626 & \textcolor{blue}{-0.24} \\
AHDMIL       & 0.838 & \textcolor{blue}{-0.03} \\
R$^2$T-MIL       & 1.062 & \textcolor{red}{+0.20} \\
MFC-MIL       & 1.506 & \textcolor{red}{+0.64} \\
IIB-MIL       & 1.842 & \textcolor{red}{+0.98} \\
\midrule
LadderMIL (Ours) & 0.866       & Ref.         \\                      
\bottomrule
\end{tabular}
}
\end{table}

\section{Interpretability}
\begin{figure}[h]
\centering
\begin{center}
\centerline{\includegraphics[width=0.48\textwidth]{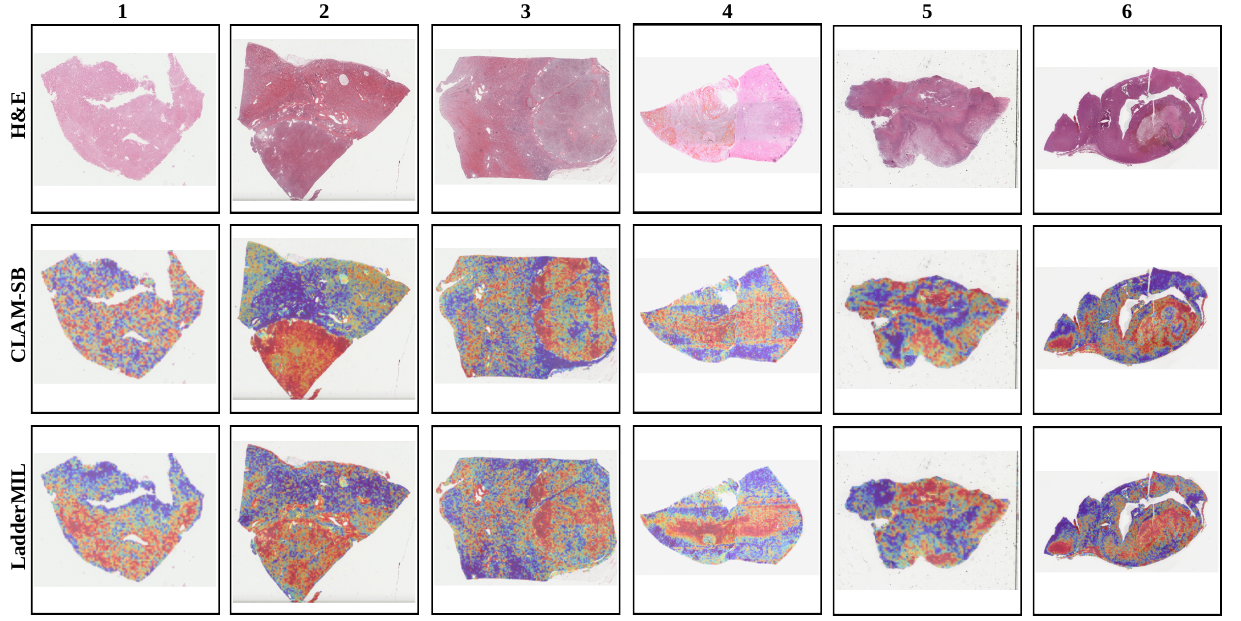}}
\caption{\textbf{Heatmap comparison for CLAM-SB and LadderMIL on TCGA-RCC.} Both frameworks successfully focus on tumour areas, while LadderMIL captures tumour in higher resolution, due to the instance-level learning.}
\label{fig:heatmap_rcc}
\end{center}
\vskip -0.4in
\end{figure}

To assess the interpretability, we visualised the attention heatmaps for CLAM-SB and LadderMIL on TCGA-RCC and our internal breast cancer cohort. Heatmaps were examined by a Consultant Histopathologist (TJK) with 25 years of diagnostic surgical pathology experience; although such assessment is necessarily subjective, expert subjective evaluation remains the basis of clinical histopathology and is the basis of ground-truth annotations for computational pathology studies. In the TCGA-RCC subtyping task (Figure~\ref{fig:heatmap_rcc}), both frameworks generally focus on tiles containing tumour, as expected. In some areas, CLAM-SB is 'out-of-focus' (\ie,\ column 1) such that areas of high attention contain fewer tumour cells mixed with more abundant stroma. Column 2 also shows that CLAM-SB tends to focus on large areas of tumour while neglecting scattered tumour cells. In contrast, LadderMIL successfully more accurately focuses on tumour cells in higher resolution, due to its instance-level learning ability that helps in discriminating single patches containing tumour.

\begin{figure*}[h]
\centering
\vskip -0.2in
\begin{center}
\centerline{\includegraphics[width=0.8\textwidth]{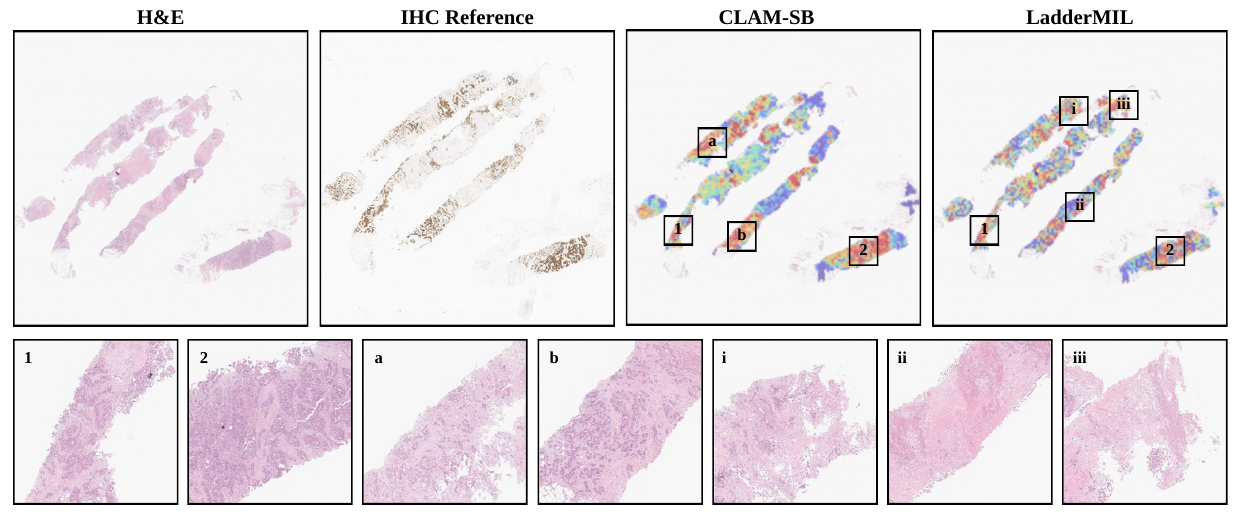}}
\caption{\textbf{Heatmap analysis for CLAM-SB and LadderMIL on an ER+ example that both model successfully classified.} (1,2) Tumour area that both model considers important. (a,b) Tumour area that CLAM-SB highlights, while LadderMIL not paying attention to. (i,ii,iii) LadderMIL also considers tumour-related stroma and inflammatory cell infiltration.}
\label{fig:heatmap_er_pos}
\end{center}
\vskip -0.3in
\end{figure*}

By analysing the results of ER status classification, we show that both CLAM-SB and LadderMIL can classify positive cases, while LadderMIL is more accurate at the classification of the ER negative cases. To understand this behaviour, we compared the ER status classification heatmaps with the immunohistochemistry (IHC) reference\footnote{IHC is the clinical gold standard for determining receptor status that uses antibody staining to detect antigens in tissue samples~\cite{walker2008ihc}.}, where strong brown nucleus staining indicates ER+ cells. Figure~\ref{fig:heatmap_er_pos} shows an ER+ example that both frameworks classified successfully. By comparing the heatmaps with the IHC reference, we find that regions of high attention align closely with areas of IHC-positive. However, in detail, CLAM-SB purely focuses on tumour cells (\ie,\ 1, 2, a, b), while LadderMIL not only focuses on tumour cells (\ie,\ 1, 2), but also focuses on stroma and inflammatory cells (\ie,\ i, ii, iii), which suggests information being drawn from the tumour microenvironment. 

\begin{figure*}[h]
\centering
\begin{center}
\centerline{\includegraphics[width=0.8\textwidth]{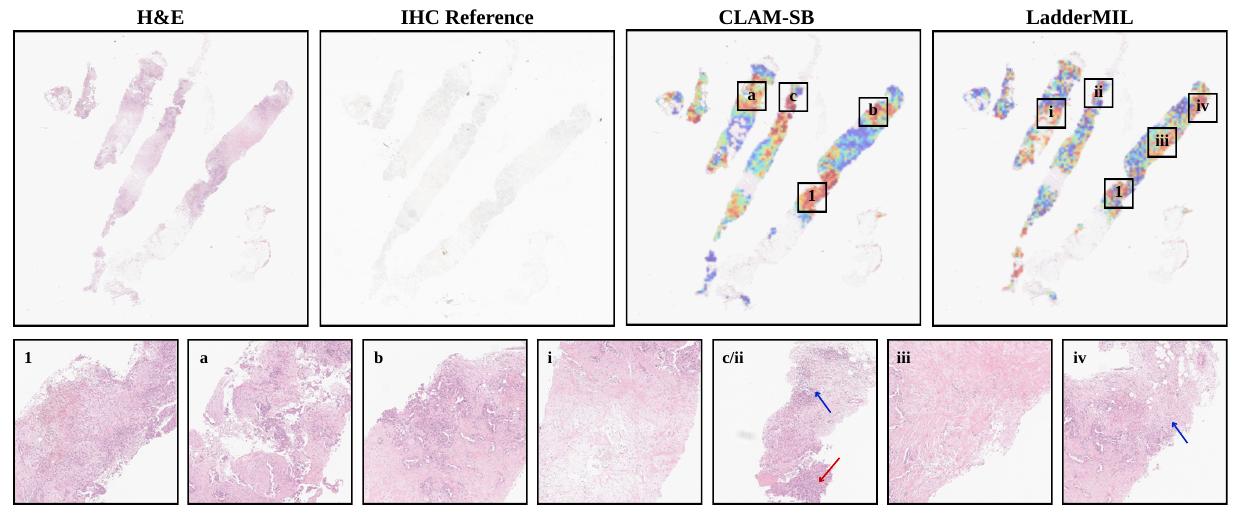}}
\caption{\textbf{Heatmap analysis for CLAM-SB and LadderMIL on an ER- example that CLAM-SB failed to classify while LadderMIL succeeded.} (1) Tumour area that both model considers important. (a,b) Tumour area that CLAM-SB highlights, while LadderMIL not paying attention to. (i,ii,iii) LadderMIL also considers tumour-related stroma and inflammatory cell infiltration. (c/ii) In merely the same region, CLAM-SB sticks to highlighting the tumour cells, while LadderMIL focuses on inflammatory cells. Red arrow points to tumour, while blue arrow points to inflammatory cells.}
\label{fig:heatmap_er_neg}
\end{center}
\vskip -0.3in
\end{figure*}

To further understand the reason why LadderMIL performs better, we studied an ER- example that CLAM-SB failed to classify correctly but that LadderMIL succeeded in Figure~\ref{fig:heatmap_er_neg}. CLAM-SB was able to consistently focus on tumour cells but failed to determine whether these cells are ER positive or negative. In contrast, LadderMIL not only paid attention to tumour areas (\ie,\ 1) but also highlighted stroma (\ie,\ i, iii) and inflammatory cells (\ie,\ ii, iv), which implies the information from other cells and 'host' areas of the tumour, especially tumour-related stroma and inflammatory cells, could help with the better classification. It is clinically reasonable for such a finding, since the interaction between malignant tumour cells and 'host' non-malignant cells in the tumour microenvironment is bidirectional and central to tumour behaviour~\cite{almagro2022tme, zhao2023tme, mo2024tme}. We attribute this improvement to the design of CFSD that enables the discrimination of each instance and the CEG that encodes instance-level information with two-dimensional coordinates to form contextual encoding at the broader slide-level. The ability to capture inflammatory cells is potentially clinically relevant as clinical research has shown that high stromal tumour-infiltrating CD8+ lymphocytes are a negative marker for therapeutic response in ER+ breast cancer~\cite{kinsella2026til}. Simultaneously, learning from tumour-infiltrating lymphocytes may also identify cases with the improved treatment response and prognosis associated with triple-negative breast cancer (TNBC) and HER2-positive breast cancer~\cite{denkert2018til}. The heatmap analysis suggests the classification decisions of LadderMIL are based on capturing relevant biological features, and are clinically interpretable and insightful.

\section{Conclusion}
In this paper, we propose LadderMIL, a novel framework that integrates the coarse-to-fine self-distillation (CFSD) paradigm and the contextual encoding generator (CEG) for multiple instance learning (MIL). CFSD enables efficient instance-level supervision by probing and distilling a classifier trained with bag-level labels, thereby addressing the limited instance-level learnability of MIL in a self-improving manner. CEG mitigates issues arising from the discontinuity of instances in background-removed WSIs and enhances the use of inter-instance contextual information by encoding precise coordinates and the bag-level attention map. The overall framework aligns with the decision-making and reasoning processes of pathologists, who assess both bag-level and instance-level features in parallel, and its ability to capture tumour surrounding tissue can potentially contribute to challenging computational pathology tasks, such as receptor status classification, treatment outcome prediction, and prognosis prediction which requires not only focus on tumour but to capture wider tumour heterogeneity. By incorporating CFSD and CEG, LadderMIL outperforms state-of-the-art frameworks, demonstrates instance-level learnability, and provides clinically reasonable interpretability.

\appendices

\section{Lemma and Condition for Proving Instance-level Learnability~\cite{jang2024learnability}}
\label{appendix:prove_inst_learnability}
Given $\mathcal{H}$ denotes the hypothesis space, $\mathcal{H}_{inst_i}$ is the $i^{th}$ instance hypothesis space, where $\mathcal{H}_{inst_i}=\{h_i:h_i(X_i)\rightarrow Y_i\}$. And $\mathcal{H}_{add_i}$ is the extra hypothesis space from external values for the $i^{th}$ instance. With $\mathcal{X}:=\{\mathcal{X}_{inst_1},\mathcal{X}_{inst_2},...,\mathcal{X}_{inst_N}\}$ to be the bag-level feature space and $\mathcal{Y}:=\{1,...k\}$ to be the bag label space, we have:
\begin{condition}$\mathcal{H}_{add_i}$ must be a subset of $\mathcal{H}_{inst_i}$ that:\\
\begin{equation}
    \mathcal{H}_{inst_i}\supset\mathcal{H}_{add_i}:=\{h_{add_i}: \mathcal{X}_{add_i}\rightarrow\mathcal{Y}\}
\end{equation}
\label{con:inst_learnability}
\end{condition}

\begin{lemma}Condition \ref{con:inst_learnability} is a necessary condition for the learnability of instances, when the hypothesis space for the $i^{th}$ instance of a MIL algorithm is $\mathcal{H}_{inst_i}\cup \mathcal{H}_{add_i}$, where $\mathcal{H}_{inst_i}$ denotes the hypothesis space for the $i^{th}$ instance and $\mathcal{H}_{add_i}$ denotes the hypothesis space for the $i^{th}$ instance generated through elements outside the $i^{th}$ instance.
\label{lemma:inst_learnability}
\end{lemma}

\section{Annotation Protocol for Receptor Status Classification}
\label{appendix:receptor_annotation}
In clinical practice, both ER and PR are scored using a proportion score ($PS$) and an intensity score ($IS$), where $PS\in \mathbb{Z}\cap[0,5]$ and $IS\in \mathbb{Z}\cap[0,3]$. These scores are then combined to form a total score ($TS$), where $TS\in \mathbb{Z}\cap [0,8]$, with $TS\ne1$, and a higher score indicates greater receptor positivity. When converting into binary positive or negative status for classification, we take $TS$ of 0 and 2 as negative, and $TS$ from 3 to 8 as positive, in line with the clinical guideline~\cite{iccr_allred}. Note that a $TS=1$ does not exist, as either $PS = 0$ or $IS = 0$ would imply the absence of receptor expression.

% \section*{Acknowledgment}

% The preferred spelling of the word ``acknowledgment'' in American English is 
% without an ``e'' after the ``g.'' Use the singular heading even if you have 
% many acknowledgments. Avoid expressions such as ``One of us (S.B.A.) would 
% like to thank $\ldots$ .'' Instead, write ``F. A. Author thanks $\ldots$ .'' In most 
% cases, sponsor and financial support acknowledgments are placed in the 
% unnumbered footnote on the first page, not here.

\section*{References}
% \vskip -0.2in
\bibliographystyle{ieeetr}
\bibliography{reference}

\end{document}